# A Review on Text-Based Emotion Detection - Techniques, Applications, Datasets, and Future Directions

Sheetal Kusal[1] · Shruti Patil[2] · Jyoti Choudrie[3] · Ketan Kotecha[2] · Deepali Vora[1] · Ilias Pappas[4]

**Abstract**

AI has been used for processing data to make decisions, interact with humans, and understand their feelings and emotions. With the advent of the internet, people share and express their thoughts on day-to-day activities and global and local events through text messaging applications. Hence, it is essential for machines to understand emotions in opinions, feedback, and textual dialogues to provide emotionally aware responses to users in today's online world. The field of text-based emotion detection (TBED) is advancing to provide automated solutions to various applications, such as businesses, and finances, to name a few. TBED has gained a lot of attention in recent times. The paper presents a systematic literature review of the existing literature published between 2005 to 2021 in TBED. This review has meticulously examined 63 research papers from IEEE, Science Direct, Scopus, and Web of Science databases to address four primary research questions. It also reviews the different applications of TBED across various research domains and highlights its use. An overview of various emotion models, techniques, feature extraction methods, datasets, and research challenges with future directions has also been represented.

[1] Symbiosis Institute of Technology, Symbiosis International (Deemed University), Pune, Maharashtra, India.

[2] Symbiosis Centre for Applied Artificial Intelligence (SCAAI), Symbiosis Institute of Technology, Symbiosis International (Deemed University), Pune, Maharashtra, India

[3] University of Hertfordshire, Hatfield, Hertfordshire, United Kingdom,

[4] University of Agder, Norway.

# 1 Introduction

Artificial Intelligence (AI) is the part of computer science that focuses on designing intelligent computer systems that show the traits we re-late with human intelligence like comprehending languages, learning problem-solving, decision making, etc. One of the significant contributions of AI has remained in Natural Language Processing (NLP), which glued together linguistic and computational techniques to assist computers in understanding human languages and facilitating human-computer interaction. Machine Translation, Chatbots or Conversational Agents, Speech Recognition, Sentiment Analysis, Text summarization, etc., fall under the active research areas in the domain of NLP. However, in the past few years, Sentiment analysis has become a demanding realm. Nowadays, Artificial Intelligence has spread its wings into Thinking Artificial Intelligence and Feeling Artificial Intelligence (Huang and Rust 2021). Figure 1 shows the sub domain of artificial intelligence. Thinking AI is de-signed to process information in order to arrive at new conclusions or decisions. The data are usually unstructured. Text mining, speech recognition, and face detection are all examples of how thinking AI can identify patterns and regularities in data. Machine learning and deep learning are some of the recent approaches to how thinking AI processes data. Existing decision-making applications include IBM Watson - a question-answering computer system, recommender systems, and expert systems. Feeling AI is developed for two-way human-machine interactions as well as the analysis of human feelings and emotions. Sentiment analysis, text-to-speech technology, chatbots that mimic human dialogue, and robots with special technology for detecting emotional signals are some of the recent technologies in feeling AI. "Feeling AI" refers to artificial intelligence (AI) that measures, understand, simulates, and reacts to human emotions. Sentiment Analysis is the area that comes un-der the umbrella of feeling AI. Emotion Detection, the adjunct of Sentiment Analysis, is intended to extract and analyze fine-grained emotions such as happiness, anger, sadness, etc. In contrast, Sentiment analysis is targeted to interpret opinions and give polarities like positive, negative, or neutral by understanding the feelings (Francisca et al. 2020). Emotions have remained important in Human-Computer Interaction and enable computer systems to understand human conduct or be more emotionally aware systems. So, technology-driven by emotions plays a critical role in decision-taking, which could be helpful to the broader range of application domains. This more comprehensive range of application domains includes Management and Marketing, User Interaction, Healthcare, Education, Finance, Public Monitoring, etc.

Owing to the evolution of technology and the tremendous growth of the Internet, massive amounts of digitized data in text, images, videos, etc., are available on digital social platforms. Digitized text is available on digital social media in blogs, news articles, customer reviews on different products, services, and their experiences, discussion forums, review/recommendation systems, conversational agents, etc. Moreover, these online social platforms permit people to express their ideas, opinions. So, all these social platforms lead the way to analyze the people's emotions, feelings for investigating social trends, tracing the customer feedback to understand and develop the business strategy, helping customers/consumers in the decision-making process, etc. Therefore text-based emotion detection has transformed into a significant research area.

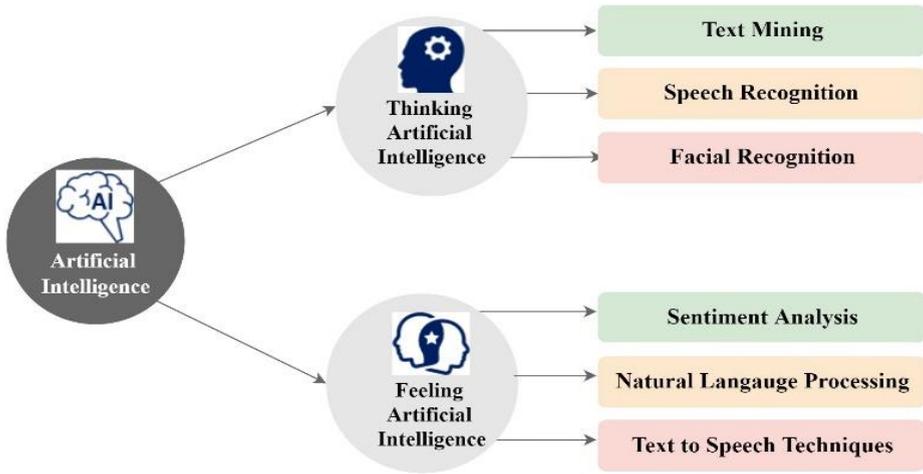

**Fig. 1** Subdomains of Artificial Intelligence (AI)

Since then, text-based emotion detection has been clasped in many real-world applications; (Chung et al. 2018) describes a novel system for emotion analytics that uses social media content to aid information collection, situational awareness, and administrative decision-making. (Sameh et al. 2020) explored automatic sentiment analysis for assessing the polarity of a product or a service review. (Winster et al. 2020) discussed emotion classification in News Articles. (Chehal et al. 2020) implemented a system based on user reviews in e-commerce recommendations. (Graterol et al. 2021) propose a framework to allow social robots to detect emotions. (Moisés R. et al. 2020) surveyed intelligent conversational agents and stated agents potentially detect neuropsychiatric diseases. (Rodrigo Bavaresco et al 2021) focused on human-like conversational agents for business. (Montenegro et al. 2019) explored conversational agents in health care. (Topal et al. 2016) developed a movie review system to help users in decision-making.

Detecting emotions from a massive amount of digitized text on social platforms to enhance man-machine interaction and aid decision-making in the various application domains is the key to encouraging research in the field. TBED has multidisciplinary facets thus applied to multiple areas such as psychology, sociology, human-computer interaction (HCI), data mining. This Systematic Literature Review (SLR) not only bring forth an outline of the current state of research on text-based emotion detection but underline the various digital social platforms that can be used for the purpose, what techniques have been applied, what different datasets have been used in the research and what will be future directions in the field.

## 1.1 Significance

The importance of emotion detection from text is rising expeditiously hand in hand with the Internet and online digital media. Therefore, technologists, business strategists, government agencies, political analysts accentuated and want to take advantage of this field in all aspects of the decision-making to improve businesses, reputations, etc. The Internet offers online social relationship sites such as Facebook, media sharing networks like YouTube, Instagram, microblogging sites like Twitter, Reddit. This increase in the advent and popularity of social networks has motivated researchers to investigate online content and analyze users' online social behaviors (Bazzaz et al. 2020).

Customer or product reviews play an essential role in determining consumer behaviors and mindsets for businesses and simultaneously helping consumers decide to buy products/services (Sunil et al. 2021). So, in addition, the adaption of chatbots or conversational agents in fields like in healthcare to monitor and automate the delivery of healthcare services (Montenegro et al. 2019), in business to improve the service quality, market competitiveness (Bavaresco et al. 2020), in education to promote meaningful interaction in distance learning or E-learning (Donggil et al 2017). So, it becomes vital to study emotions from the text in different application domains.

## 1.2 Motivation

There is no existing Systematic Literature Review (SLR) that emphasizes techniques, datasets, application domains, comparative analysis, and future directions in text-based emotion detection. Present reviews and surveys lack a comprehensive study of applications domains and future trends in text-based emotion detection. The most common way for humans to interact with computers is still text-based input; even in the advent of web 3.0, text-based emotion detection should be addressed as a research area. There has been a considerable surge in the field of sentiment analysis and emotion detection in the last few years. Although the present paper focuses on the research articles on text-based emotion detection, many researchers have focused on sentiment analysis. As per our extensive study, no systematic reviews have not been performed on text-based emotion detection with significance on techniques, datasets, applications domains, and future directions.

This systematic review intends to show insufficiency in the literature on the subject of datasets. Even though the alike datasets are available, each dataset has a limited amount of data, and the labels related to emotions are limited; as a result, emotion detection systems cannot be reliably trained or generalized (Feng et al. 2020).

Another inspiration for this systematic review is that text-based emotion is highly domain-dependent. Thus, approaches and datasets suitable in one domain like social media may not be ideal for another application domain like review systems (Loitongbam et al. 2020). So, such topics need to be studied to identify the research gaps in the field of TBED. The essence of this SLR is to highlight existing approaches, available datasets, applications, challenges, and future directions in text-based emotion detection.

## 1.3 Terms and terminologies

The terms (Tyng et al. 2017) that are typically used in text-based emotion detection research are as follows:

1) Sentiment - Sentiment is defined in the Oxford dictionary as exaggerated and self-indulgent feelings of tenderness, sadness, or nostalgia or a feeling or an opinion, especially one based on emotions.

2) Emotion - Emotions represents a composite set of interactions between objectives and subjective variable quantities that are facilitated by the brain and hormonal systems, which can -

(a) escalate the affective experiences of emotional valence like displeasure and pleasure and emotional arousals like low-high activation or arousing-calming.

(b) produce logical processes which are relevant to emotions such as appraisals, perceptual affect, labeling.

(c) cause extensive physiological and psychological changes in response to the stimulating circumstances.

(d) encourage communicative, goal-directed, and adaptive conduct, albeit this isn't always the case.
3) Mood - Moods remain retentive than emotions, which has both negative and positive characteristics.
4) Feelings - Feelings are cognitive perceptions that are valanced, either positive or negative, and are followed by the body's physiological changes in the body, particularly the internal organs, in order to maintain or restore homeostatic balances.
5) Affect - Affect is a term that relates to consciously reachable basic processes such as arousal and pleasure. It is difficult to explain individuals experienced emotional feelings but has been connected to physical states such as homeostatic drives, e.g., thirst and hunger, and external stimuli like touch, taste, smell, auditory, visual.

## 1.4 Evolution of text-based emotion detection

The term "Emotion" was inscribed in the work "The Expression of the Emotions in Man and Animals" in 1872, by the famous Scientist, Charles Darwin. It was the first theory where human emotion and expressions were identified. Then in 1980, Psychologist R. Plutchik (Plutchik 1980) built the emotion wheel. He showed eight basic and principal emotions: anticipation, anger, fear, disgust, joy, trust, surprise and sadness.

He further said that each fundamental emotion has a polar opposite. After that, the first emotion recognition system was developed based on humans' facial expressions (Paul et al. 1980) in 1980. Paul Ekman and Wallace Friesen developed a system named EFACS (Emotion Facial Action Coding System). Paul Ekman determined people of different cultures share six fundamental emotions represented by certain facial expressions. Happiness, anger, disgust, surprise, fear, and sadness are the six primary emotions. Then research in emotion detection advanced with recognizing emotions in speech. The first seminal work on identifying emotions in the speech was written by (Dellaert et al. 1996), which became the first research publication in the field.

Figure 2 shows the brief history of emotion detection and how emotions and their research evolved. Rosalind Picard's affective computing theory, published in 1997, was the next turning point in the area. According to Picard (Picard et al. 1997), computers must have the ability to identify, interpret, and express emotions if we want them to be truly intelligent and communicate naturally with us. As a result, it shifted the field of emotion detection in a new direction. Affecting computing theory enhanced human-computer interaction through text-based inputs. In 1997, the first emotion dataset developed was ISEAR. The International Survey on Emotion Antecedents and Reactions (ISEAR) database is created from multicultural questionnaire studies from 37 countries. It consists of 7665 sentences labeled with emotions and emotion labels as anger, disgust, fear, guilt, joy, sadness, and shame.

In 1998, the first English language lexical database was created for Natural Language Processing (NLP) tasks as Wordnet (Fellaum et al. 1998). Then practical use of the term sentiment and emotion analysis was coined in 2001 by Das and Chen (Das and Chen 2001) to predict the stock market. After that, Wiebe (Wiebe et al. 2005) wrote the first article on annotating emotion and opinion detection from the text in 2005. After that, in 2009, SenticNet, is publicly available semantic resource for concept-level sentiment analysis, was developed. Word embedding is a term used to represent words for text analysis in NLP. So, the first-word embeddings Word2Vec were developed in 2013. Then in the same year as 2013, neural networks were adopted for the first time in the NLP task. Then, in 2017, the concept of transformers was explained in the publication (Vaswani et al. 2017), which gave a huge surge in the field.

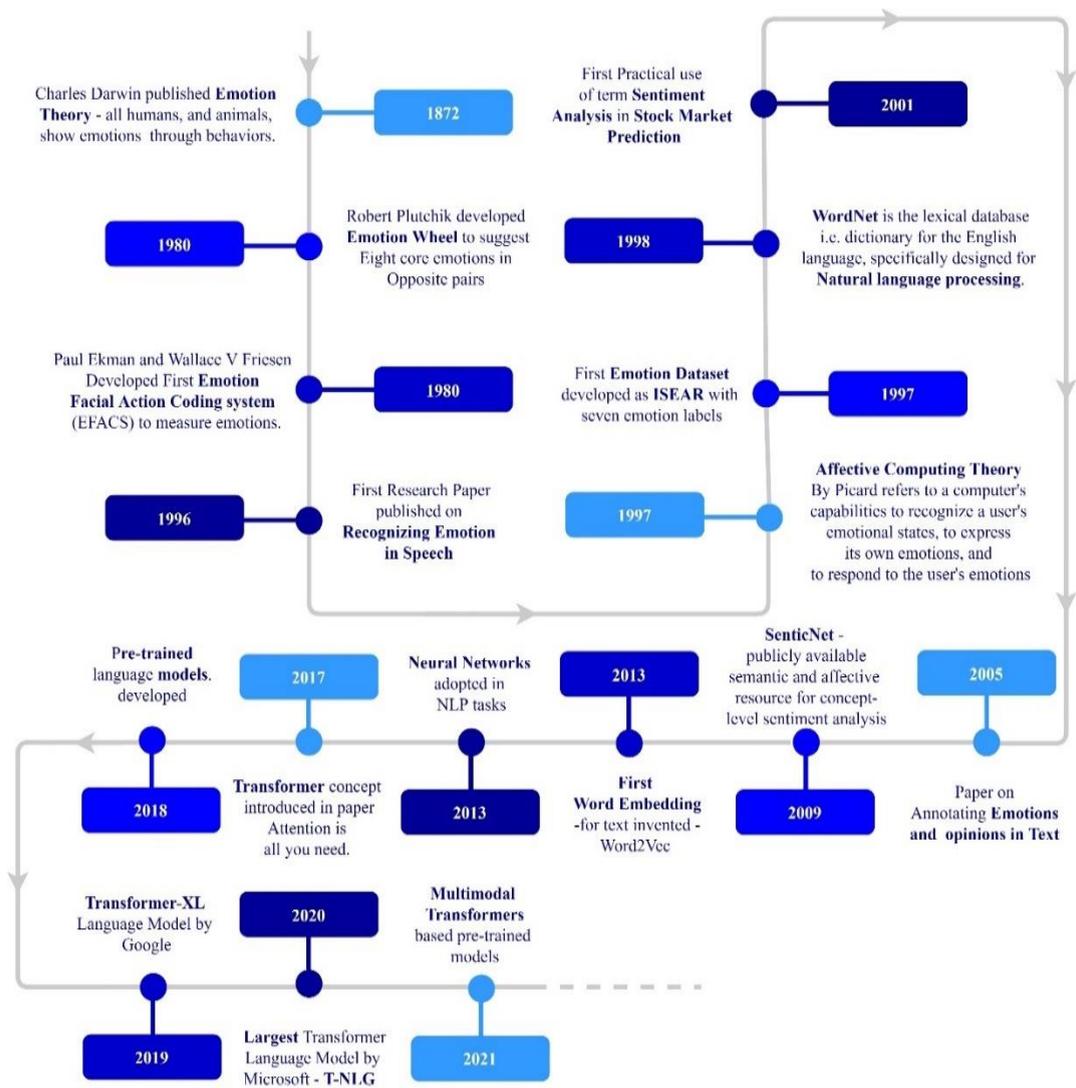

**Fig. 2** Evolution of emotion detection – A brief history of emotion detection.

And then pretrained model brought an evolution in the field of automated text classification and analysis in 2018 in the form of BERT. In 2019, Google developed the Transformer XL model to help machines understand context beyond that fixed-length limitation. Microsoft introduced a transformer-based generative language model - Turing Natural Language Generation (T-NLG) in 2020. It is the largest model published with 17 billion parameters. In 2021, the research field directed towards integrating multimodal information using large pretrained models. We look at new techniques and methods evolving in the text-based emotion detection field in the coming sections.

## 1.5 Prior Research

On the topic of text-based emotion detection, inconsiderable literature reviews have been written. As per our best knowledge, no systematic literature review has been written so far on text-based emotion detection. Indeed, emotion detection evolved from sentiment analysis. Hence, it becomes sensible to consider systematic reviews on sentiment analysis in this survey. Table 2 gives an overview of survey papers. In a tertiary analysis, (Ligthart et al. 2021) explored at sentiment analysis.

The authors gave a thorough overview of the various approaches to sentiment analysis. In this sentiment analysis paper, the features, methodologies, and datasets utilized are explored. Difficulties and unaddressed issues are also mentioned, which can aid in identifying areas where sentiment analysis research is needed. (Alswaidan and Menai 2020) discussed explicit and implicit emotion detection from the text in their survey paper. The authors presented the state of art approaches, their main features with merits and demerits. Moreover, the authors studied different corpora and different lexicons available for emotion detection from text. This review also underlines the significance of NLP tasks, performances of different approaches; and suggests some open issues. (Francisca et al. 2020) surveys the concept and main approaches implemented in TBED systems. Authors added discussion on futuristic approaches in the field with datasets, outcomes with positives, and negatives.

Also, annotated datasets are provided with suitable text for emotion detection. This article, too, presents a few open problems and future research directions. (Akshi Kumar et al. 2019) aims a systematic literature review to investigate and review current work on sentiment analysis based on context and highlights research gaps and future directions. This SLR identified and analyzed the use of context in Sentiment Analysis. (Seyeditabari et al. 2018) reviews the work that has been done in TBED. It argues that many techniques, methodologies, and models developed to detect emotion in text, are inadequate for various reasons. (Shervin et al. 2021) presents a thorough overview of deep learning-based text classification models, including their methodological contributions, similarities, and weaknesses. (Soujanya et al. 2017) reviews affective computing, multimodal affect analysis frameworks. (Bostan et al. 2018) focuses mainly on the use of audio, visual, and text information and studies the fusion techniques of multimodal data. (Dang et al. 2020) reviews the methods of deep learning employed to resolve sentiment analysis problems, like sentiment polarity and comparative analysis of different deep learning methods. (Shilpi et al. 2017) reviews different approaches, classifiers, applications domains, and issues related to text emotion detection. (Koswari et al. 2019) provides an overview of text classification algorithms, various feature extraction approaches, existing algorithms, dimensionality reduction methods, methodologies, and evaluation methods. It also covers the techniques' limitations and real-world implementations. (Ali et al. 2017) presents a classification of sentiment analysis, a survey on emotion theories, emotion mining-related polarity classification methods, and resources, useful resources, such as datasets and lexicons. (Sailunoz et al. 2018) discusses advancements in emotion detection research, including several emotion models, related datasets, detection algorithms, characteristics, limits, and prospective future approaches in the field of text and speech-based emotions.

Presented surveys ((Francisca et al. 2020, Ligthart et al. 2021, Alswaidan and Menai 2020, Akshi et al. 2019, Seyeditabari et al. 2018, Shervin et al. 2021, Soujanya et al. 2017, Bostan et al. 2018, Dang et al. 2020, Shilpi et al. 2017, Koswari et al. 2019, Ali et al. 2017, Sailunoz et al. 2018) lack comprehensive study of datasets, technical approaches, a study of application domains, and future directions. This review initiates prospects for future research work in this area. It also aims at an in-depth survey on various datasets, techniques, approaches, application domains or areas, and future trends in the field of TBED.

## 1.6 Research Goals

The research goal of this systematic literature review (SLR) is to gain a comprehensive understanding of the current progress in TBED and recognize open challenges and barriers in the area of TBED. Table 1 shows research goals.

**Table 1** Research Questions with objectives

| RQ Number | Research Question | Objective |
| --- | --- | --- |
| RQ1 | What are the different Artificial Intelligence (AI) approaches used for TBED? | The aim is to study the different Artificial Intelligence (AI) techniques for TBED, their advantages, and limitations and to show a comparative analysis of different techniques. |
| RQ2 | What application domains have been adopted in text-based emotion detection? | Different application domains have different requirements and, therefore, require different methods, datasets. The aim is to study the different application domains in text-based emotion detection, their requirements, and comparative analysis. |
| RQ3 | What are the different datasets available for research purposes in text-based emotion detection, and which domains have been acquainted in the available data sets? | The goal is to review the available public datasets for text-based emotion detection by analyzing application domains, data sources, data size, emotion labels, and data imbalance. |
| RQ4 | What are the difficulties and open issues concerning TBED? | Text-based emotion detection is a domain-dependent task. Therefore, it is one of the challenges. Therefore, the goal is to identify the challenges and open issues in text-based emotion detection. |

## 1.7 Contribution of The Work

The Contributions of this systematic literature review are as highlighted:
- We provide a detailed overview of the existing literature on text-based emotion detection with a particular focus on techniques, datasets, applications, associated challenges, and future directions.
- We discuss and try to explore the techniques and methods used and how they are refurbishing the research in text-based emotion detection.
- We also provide an overview of different publicly available datasets that can support research in the domain.
- We also focus on analyzing the role of text-based emotion detection in various application domains and explore each application domain.
- We present different types of challenges associated with text-based emotion detection, including dataset, the accuracy of existing techniques, handling quality of data.

Figure 3 shows the outline of the SLR. It shows sections, subsections, and discussions aimed at the specific sections.

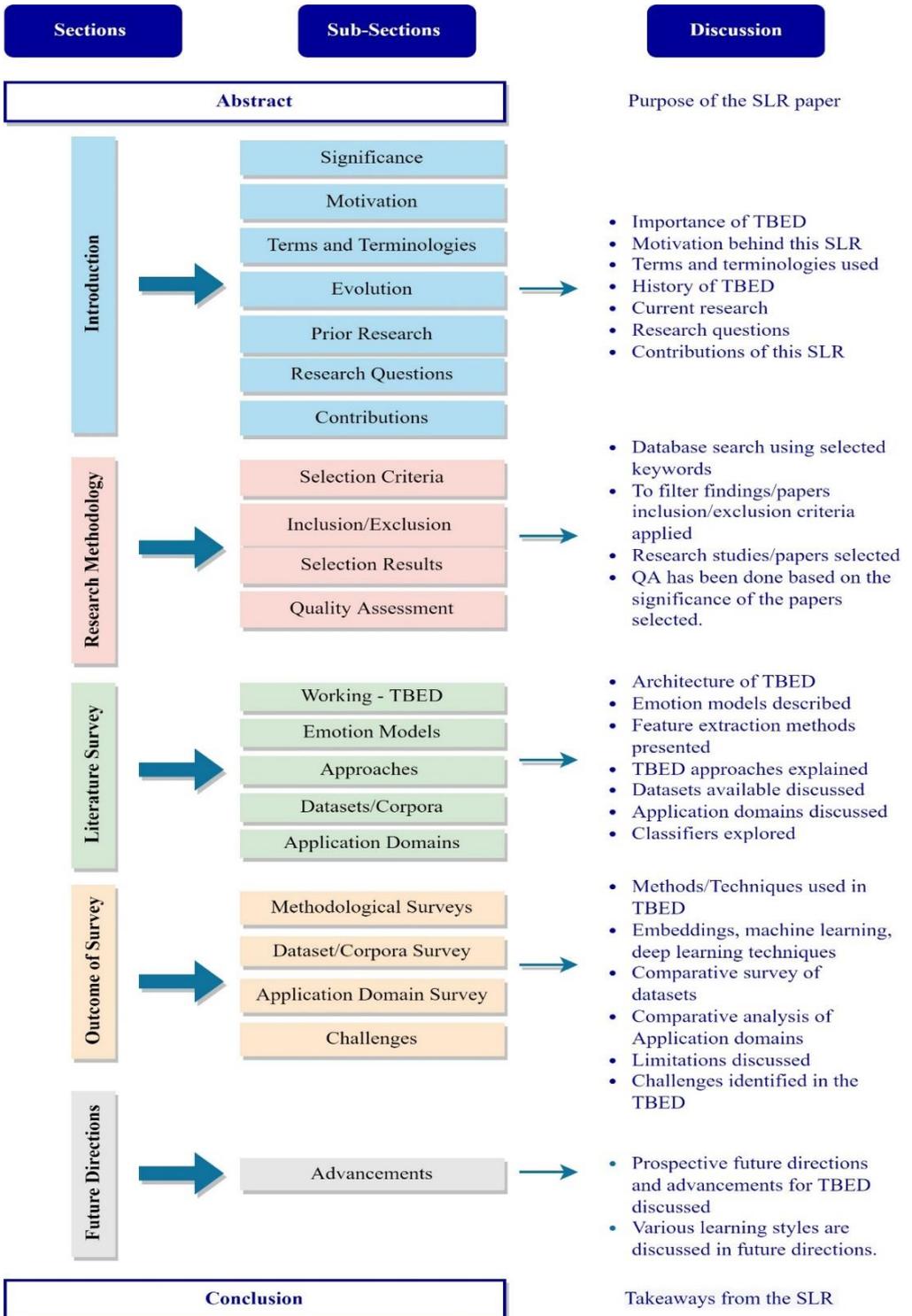

**Fig 3** Outline of the SLR

# 2 RESEARCH METHODOLOGY

An SLR is a type of ancillary research that practices a definite process to discover, analyze, and understand existing information relevant to a certain research issue in a fair and (to some extent) reproducible manner (Kitchenham et al. 2007). This systematic literature review (SLR) follows the PRISMA guidelines. The Preferred Reporting Items for Systematic Reviews and Meta-Analysis (PRISMA) method is a collection of guidelines for the composition and structure of systematic reviews and other meta-analyses based on data. These guidelines were published by (Moher et al. 2009) in 2009. Selection criteria, inclusion/exclusion criteria, selection outcomes, and quality assessment are the four sections of the authors' research methodology.

## 2.1 Selection Criteria

Authors predominantly employed IEEE, Science Direct, Scopus, and Web of Science databases for searching the documents related to emotion detection. First, a specialized query was constructed to get the relevant articles from multiple database searches. Then, the filtering process is applied to the article selection to improve the results that meet our primary goals. This process includes eliminating duplicates, applying inclusion and exclusion criteria, filtering based on title and abstract, and finally, full-text screening. Figure 4 shows the process of literature review in detail. Table 3 shows search keywords or queries used for selecting the data from different databases using primary and secondary keywords using AND, OR Boolean operators.

## 2.2 Inclusion/Exclusion Criteria

Authors prepared a set of inclusion criteria for research article selection and exclusion criteria for research article rejection to choose relevant research studies for systematic review. After identifying all relevant articles, the authors eliminated the articles that did not meet the criteria by using exclusion criteria. Table 4 shows the inclusion and exclusion criteria.

## 2.3 Selection Results

With the initial query result, a total of 1782 articles were discovered. IEEE provided 584; Science Direct supplied 886, SCOPUS found 145, Web of Science gave 167 papers, articles, and research publications linked to Text-based emotion detection. Following that, 1133 articles were obtained as a result of applying inclusion/exclusion criteria. Further search narrowed down by studying keyword relevance, title, and abstract of articles; 266 articles were found. Finally, full-text filtering and quality assessment criteria were applied to each item, and 63 articles were chosen for the final literature review. From 2005 through 2021, studies from conferences, journals, and reviews were included.

Table 2 Summary of existing surveys related to text-based emotion detection

| Ref. | Data set | Emotion Models | Multi-modality | Pre-processing Techniques | Feature Extraction Techniques | Techniques /Approaches | Applications | Future Directions | Performance Metrics | TBED Challenges discussed in the survey paper | | | | Overview |
|---|---|---|---|---|---|---|---|---|---|---|---|---|---|---|
| | | | | | | | | | | Dataset | Semantic | Technique | Quality of data | |
| Francisca et al. 2020 | √ | √ | × | × | × | √ | × | √ | × | √ | √ | × | × | It surveys the concepts, main approaches, futuristic approaches, datasets used, performances, strengths, and weaknesses in the area of TBED systems. |
| Ligthart et al. 2021 | × | × | √ | √ | √ | √ | √ | √ | × | √ | × | √ | × | It presents a tertiary study that gives current research in sentiment analysis. It covers only secondary studies. It covers key points and different approaches, algorithms, datasets, the preferred deep learning models for sentiment analysis. |
| Alswaidan and Menai 2020 | √ | √ | √ | × | √ | √ | × | √ | √ | √ | √ | × | × | It focuses on implicit and explicit emotion recognition from text. It presents the state of art approaches, their features with merits and demerits, different corpora, and different lexicons available for text-based emotion detection. |
| Akshi et al. 2019 | × | × | √ | × | √ | √ | √ | × | × | √ | √ | × | × | It investigates and reviews current work on sentiment analysis based on context and discusses the limitations and future directions. |
| Seyeditabari et al. 2018 | √ | √ | √ | × | √ | √ | × | × | √ | √ | × | √ | × | It reviews the work that has been done in TBED. It argues that many techniques, methodologies, and models developed to detect emotion in text, are inadequate for various reasons. |
| Shervin et al. 2021 | √ | × | × | × | × | √ | √ | √ | √ | √ | × | √ | × | It provides a comprehensive review of deep learning models for text classification with contributions technical, and weaknesses. |
| Soujanya et al. 2017 | √ | √ | √ | √ | √ | √ | × | √ | × | × | √ | √ | × | It reviews affective computing, multimodal affect analysis frameworks. It focuses mainly on the use of text, visual and audio data and studies the fusion techniques of multimodal data. |
| Bostan et al. 2018 | √ | √ | √ | × | × | √ | √ | × | × | √ | × | × | × | It surveys the datasets, compares emotion corpora, and aggregates them in a standard file format with a common annotation schema. |
| Dang et al. 2020 | √ | × | × | √ | √ | √ | √ | × | × | × | √ | × | × | It reviews the methods employed in deep learning to resolve sentiment analysis problems, like polarity and comparative analysis. |
| Shilpi et al. 2017 | × | × | √ | √ | √ | √ | √ | × | × | × | × | × | √ | It reviews different approaches, classifiers, applications domains, and issues related to text emotion detection. |
| Koswari et al. 2019 | × | × | × | √ | √ | √ | √ | × | √ | × | × | × | × | provides an overview of text classification algorithms, various feature extraction approaches, existing algorithms, dimensionality reduction methods, methodologies, and evaluation methods |
| Ali et al. 2017 | √ | √ | √ | × | √ | √ | × | × | × | √ | × | × | × | It presents a survey on emotion theories and emotion mining related polarity classification methods and resources including lexicons and datasets. |
| Sailunoz et al. 2018 | √ | √ | √ | × | √ | √ | × | × | × | × | √ | × | √ | discusses advancements in emotion detection research, including several emotion models, related datasets, detection algorithms, characteristics, limits, and prospective future approaches in the field of text and speech-based emotions. |

**Table 3** Sources with search queries

| Database | Search Query | Count of Articles obtained | Count of articles after inclusion and exclusion criteria | Selection-Based On Title and Abstract | Final Selection after Quality Assessment & duplicates |
|---|---|---|---|---|---|
| IEEE | ("text emotion detection") AND ("artificial intelligence or deep learning or machine learning") OR ("Chatbots or Conversational agents or Social Media or Twitter or Facebook or Reddit or Instagram or Reviews") OR ("emotion analysis or emotion recognition or sentiment analysis") | 584 | 482 | 164 | 31 |
| Science Direct | ("text emotion detection") AND ("artificial intelligence or deep learning or machine learning") | 886 | 453 | 41 | 19 |
| Scopus | (TITLE-ABS-KEY (emotion AND detection) AND TITLE-ABS-KEY (text) AND TITLE-ABS-KEY (artificial AND intelligence OR deep AND learning OR machine AND learning) OR TITLE-ABS-KEY (sentiment AND analysis OR emotion AND analysis OR emotion AND recognition) OR TITLE-ABS-KEY (chatbots OR conversational AND agents OR social AND media OR Twitter OR Facebook OR Reddit OR Instagram OR reviews)) | 145 | 140 | 38 | 7 |
| Web of Science | ("text-based*") AND TOPIC: ("artificial intelligence" or "machine learning*" or "Deep learning*" or "natural language processing*") AND TOPIC: ("Emotion Detection*" OR "Sentiment analysis*" OR "Emotion Analysis*" OR "Emotion Recognition*" OR "Chatbots," OR "Conversational agents*" OR "Social Media*" OR "Twitter" OR "Facebook" OR" Reddit" OR "Instagram" OR "Reviews") | 167 | 58 | 23 | 6 |
| | Total Articles | 1782 | 1133 | 266 | 63 |

## 2.4 Quality Assessment

The following steps were used to assess quality:
1) Emotion detection: Research must be focused on emotion recognition, techniques, or datasets.
2) Text-based emotion detection: Research must focus on text-based emotion detection in social media and different application domains.
3) Artificial intelligence techniques: Research must focus on different artificial intelligence techniques used in text-based emotion detection.
4) Datasets: Research work also focuses on the publicly available datasets related to text-based emotion detection.
5) Classification Techniques – The research work emphasizes classification techniques used in text-based emotion detection.

**Table 4** List of inclusion and exclusion criteria

| Criteria | Inclusion criteria | Exclusion criteria |
|---|---|---|
| 1 | Publication Years of articles must be between 2005 to 2021 | Non-English Research articles |
| 2 | Article types considered Conference Paper, Article, Review, Conference Review | Duplicate research articles |
| 3 | Articles published in computer science, engineering, psychology, decision sciences, and social sciences were considered. | Research articles with non-availability of full text |
| 4 | At least one of the search terms should be met by articles. | Research articles that are not relevant/ not focused on text-based emotion detection and artificial intelligence. |
| 5 | Articles should give answers to the research questions. | Articles that do not provide information related to the research question. |

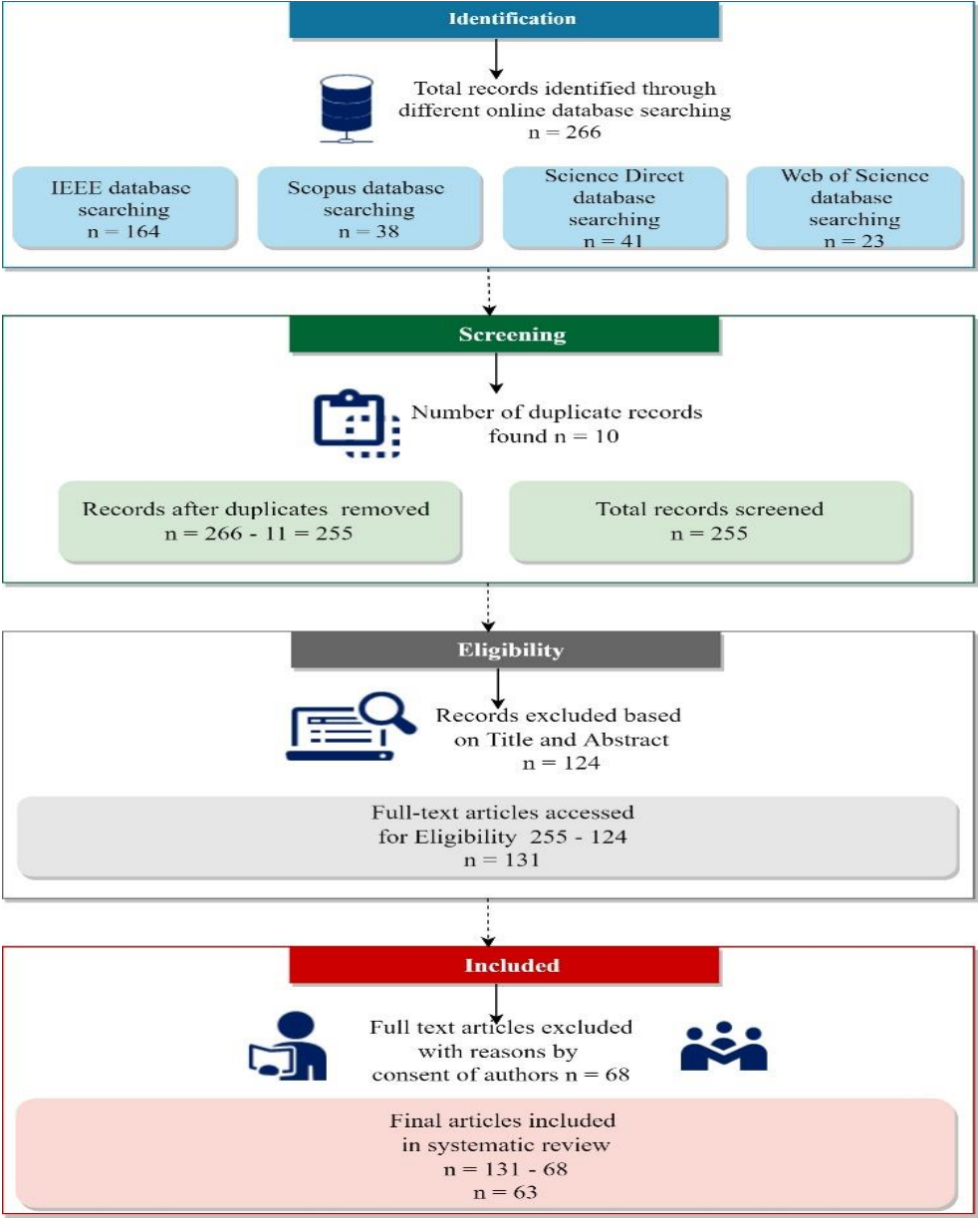

**Fig. 4** shows a detailed process of literature review

# 2 LITERATURE SURVEY

A TBED system is a set of methods that process and classify text information to detect emotions embedded in them. When we take a broad perspective, we may divide it into many categories, as shown in Figure 5. Emotion models, datasets, approaches, application domain, text pre-processing methods, feature extraction analysis, feature extraction methods, and classifiers are shown in the figure. In text-based emotion detection, online social media, product or customer reviews, conversational agents or chatbots, discussion forums are some sources that are rich in textual information. So, these application domains need to be focused on. Then there are some datasets or corpora available online which already have annotated data or labeled data. It would be useful to gain a deeper understanding of emotions to enhance the classification process. Modeling emotions has a variety of methods and approaches, and it is still ongoing. The categorical, dimensional, and componential models are preferred. The data needs to be preprocessed before features extraction. These feature extraction steps are crucial for a classification process. It reduces the original data to its most essential characteristics or makes it compatible with the classification algorithms. Text analysis generally uses five types of analysis to extract the features: lexical, syntactic, semantic, disclosure integration, and pragmatic analysis. There are different methods used for feature extraction, such as BoW, TF-IDF, POS, etc. Then all of these features are inputted into the classification system, which has a variety of classifiers at its disposal. More recently, deep learning and machine learning classifiers are commonly been used. We have given a detailed explanation of these approaches and methods in the coming sections.

## 3.1 Text-Based Emotion Detection Model

Figure 6 shows the text-based emotion detection model. In the first step, the data is acquired from different text sources. Then text preprocessing is performed to clean the data. Mainly, the text from social media, product/customer reviews consist of slang words, emojis, short text, incomplete words, etc., which requires preprocessing. So preprocessing steps include tokenization, lowercasing, removing stop words and digits, removing punctuation marks, removing white spaces, removing hashtags (#), removing URLs, removing emoticons and emojis. Then applying stemming and lemmatization to get words into their root form. The next step is feature analysis which includes feature extraction and feature selection. In Feature extraction, essential features are extracted. If the feature set is large, the feature selection can be made to choose the most informative. The final step is classification and assigning data to one of the possible emotion classes. Classification techniques are used to recognize emotions. Commonly known classification techniques are machine learning, deep learning. In the literature, we can find, for example, Support Vector Machine (SVM), K-Nearest Neighbor (K-NN), Naïve Bayes, many others. Deep learning does not require specific features. It learns from data itself. However, we need a relatively large data set to train generalization abilities. If the available data set is too small, its size can be expanded artificially by data reinforcement techniques.

### 3.1.1 Emotion Models for Emotion Detection

Studies of emotions date back to Greek mythology. Ancient Greek literature exhibits a wealth of emotions. According to Greek literature, emotions were organized into four basic categories: metus (fear), aegritudo (pain), libido (lust), and Laetitia (pleasure). After that, in the 19th century, Darwin's emotion theory stated emotions and expressions of emotions in humans and animals. In psychology, emotions are categorized into basic emotions and complex emotions (that is, emotions that are hard to classify under a single term, such as guilt, pride, shame, etc.). However, there is no universally accepted model of emotions. There are two prevalent theories in this field: the discrete emotion model and the dimensional model. And the third model has based on appraisal theory: the componential model. Table 5 shows comparative analysis of different emotion models.

1) Discrete Emotion Model – It is also termed a categorical emotion model. As stated by, discrete emotion theory, some emotions are distinguishable based on neural, physiological, behavioral, and expressive features regardless of culture. The discrete emotion model states that emotions can be placed into basic distinct classes or categories. Different emotions correspond to different neurological subsystems in the brain. Ekman's model (Ekman Paul et al. 1999) categorizes emotions into six basic categories. These basic emotions are happiness, sadness, anger, disgust, surprise, and fear. Plutchik's model (Plutchik 1980) states eight emotions in opposite pairs as joy vs. sadness, trust vs. disgust, Anger vs. fear, and surprise vs. anticipation.
2) Dimensional Emotion Model - This model assumes, emotions are dependent and a relation exists between them. The dimensional model is based on the hypothesis that all emotions result from a common and interconnected neurophysiological system. Dimensional emotion models define a few dimensions with some parameters and specify emotions according to those dimensions. Two or three dimensions are used in most dimensional emotion models— 'valence' (indicates the positivity or negativity of an emotion), 'arousal' (indicates the excitement level of emotion), and 'dominance' (indicates the level of control over an emotion). (Russell et al. 1980) presents a circular two-dimensional model prominent in dimensional emotions representation called the circumplex of effect. (Plutchik 1980) presents a 2-dimensional wheel of emotions showing valence on the vertical axis and Arousal on the horizontal axis.
3) Componential Emotion Model - This approach is an expansion of the dimensional approach. It is built on the appraisal concept (Scherer et al. 2005). According to this paradigm, emotion is derived through an analysis of circumstances if a person can be met with it. The outcome is determined by a person's background, objectives, and possibilities for action.

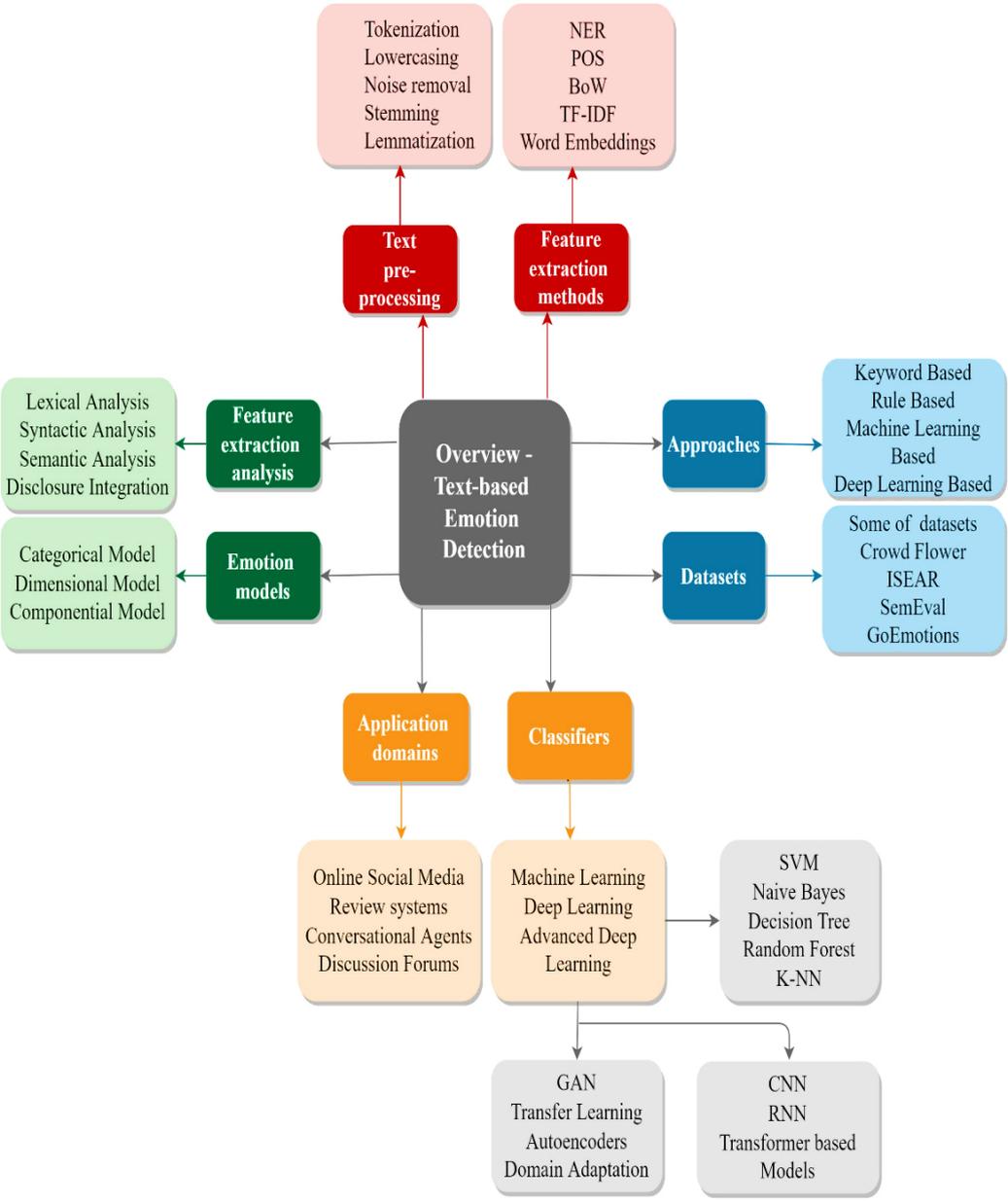

Fig. 5 Overview of TBED systems

**Table 5** Comparative analysis of different emotion models

|  | **Discrete Model** | **Dimensional Model** | **Componential Model** |
|---|---|---|---|
| **Description** | Emotions can be placed into basic distinct classes or categories. Emotions are independent. | Emotions are not independent, and that there exists a relation between them. | The componential model uses the appraisal theory based on a person's experience. |
| **Advantages** | Basic and Universal Model. Widely adopted due to its simplicity. | Complex/mixed emotions are addressed well. Highly recommended for projects involving emotional resemblances. | Focus on the variableness of different emotional states. |
| **Disadvantages** | Limited to fixed emotions. Difficult to address complex and mixed emotions | Reduction in 3-D space results in loss of information. Not all fundamental emotions are compatible with the dimensional space. | No such standard appraisal criteria. Different variants. |
| **Models** | Paul Ekman Model, Robert Plutchik Model, Orthony, Clore, and Collins (OCC) model | Russell circular two-dimensional model, Plutchik 2-D wheel of emotions, Russell and Mehrabian 3-D emotion model, The Hourglass of Emotions | Scherer's Appraisal theory. |
| **References** | Ekman Paul et al. 1999, Plutchik 1980, Ortony et al. 1990 | Plutchik 1980, Russell et al. 1980, Russell et al. 1977, Cambria et al. 2012 | Scherer et al. 2005 |
| **Emotions** | anger, fear, disgust, joy, surprise, sadness, | acceptance, anger, disgust, surprise, anticipation, fear, joy, sadness | interaction modalities (Pleasantness), interaction contents (Attention), interaction dynamics (Sensitivity), interaction benefits (Aptitude). |

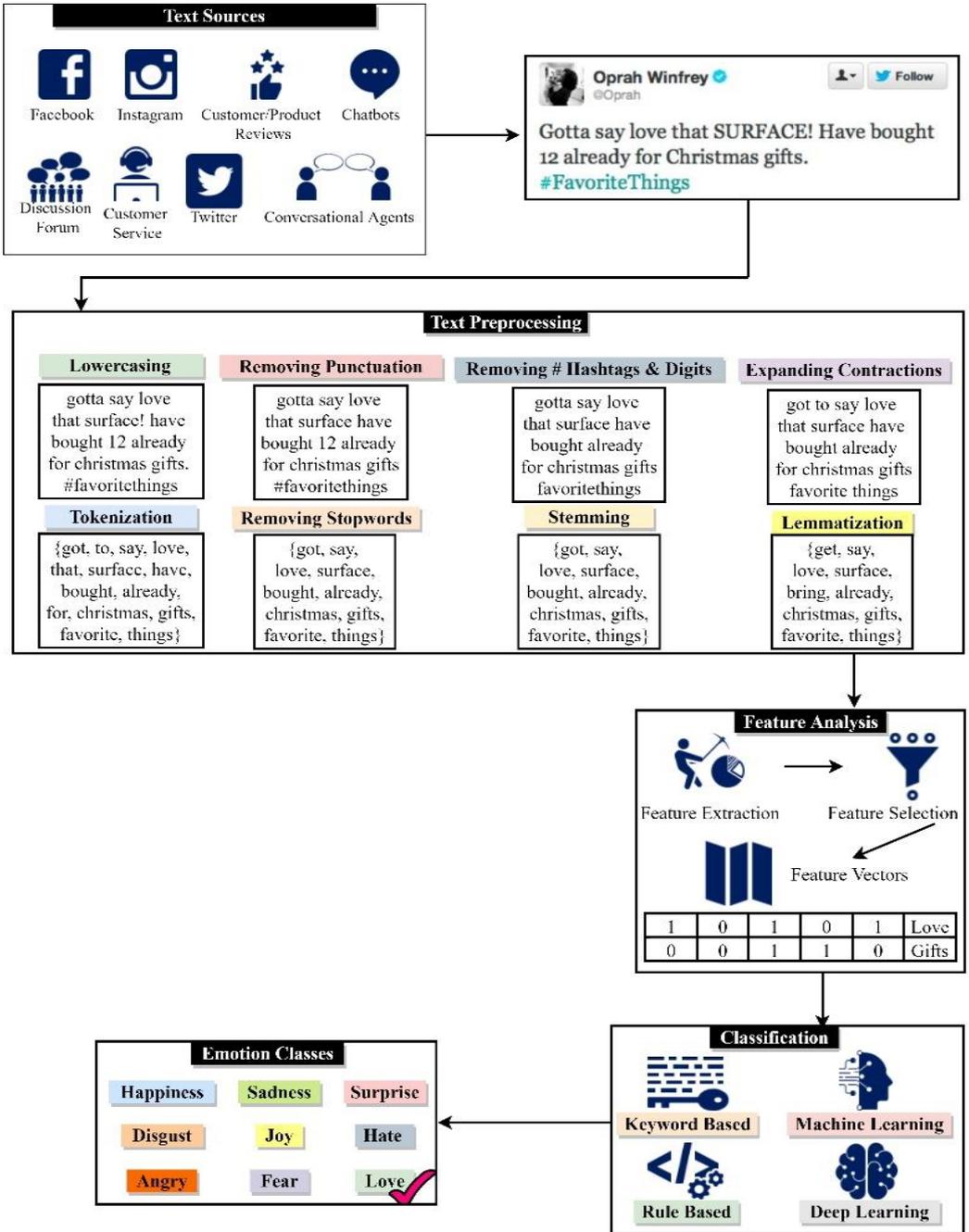

**Fig. 6** Text-based emotion detection model

### 3.1.2 Text Preprocessing Methods

The initial step in text-based emotion detection prepares the data to the form appropriate for any machine or deep learning system. Many deep learning and machine learning algorithms require numerical data to perform any classification or regression task. Therefore, text preprocessing and feature extraction are vital steps for text classification problems. This segment introduces methods of text cleaning, tokenization, normalization, and allowing for informative featurization. Mainly, the text from social media, customer/product reviews consist of short text, emojis, slang words, incomplete words, etc., which make preprocessing a necessity. Text preprocessing has the following steps:
1) Data cleaning – It includes lowercasing, punctuation, and digits removal, stop words removal, hashtag & HTML tag removal, expanding contractions, etc.
2) Tokenization/Segmentation – Splitting text/strings into tokens representing words.
3) Normalization - Stemming and lemmatization include truncating a word to its root form.

Figure 6 shows all preprocessing steps in detail with examples. In addition, the figure shows the output after applying each step.

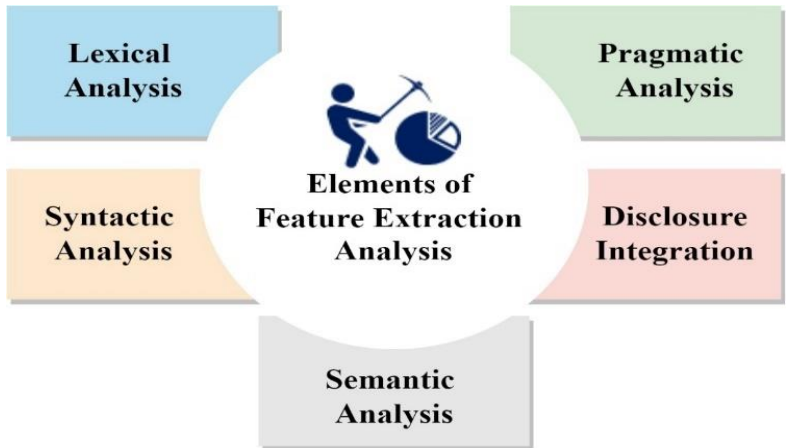

**Fig. 7** Elements of Feature Extraction Analysis

### 3.1.3 Feature Extraction Analysis

Commonly, documents and texts are unstructured data. Therefore, this section gives a brief theoretical idea about feature extraction elements before diving into feature extraction methods. Following are some elements that need to be considered while extracting features in the text. Figure 7 shows different feature extraction analyses used in text-based emotion detection.
1) Lexical Analysis – In linguistic analysis, the whole text is divided into paragraphs, sentences, and words. It helps to identify and analyze the word structures. Tokenization is one of the examples of linguistic analysis.

2) Syntactic Analysis – It involves analyzing the grammar of words in a sentence and showing the relationship among the words. Parts-of-Speech tagging (POS), parsing are some examples of syntactic analysis.
3) Semantic Analysis – Semantic analysis determines the actual meaning of words and evaluates the text's significance. Named Entity Recognition (NER) is an example of semantic analysis.
4) Disclosure Integration – Disclosure integration gives a sense of the context. It involves building the relationship and meaningfulness between two consecutive sentences or phrases. Dependency parsing is one of the examples of disclosure integration.
5) Pragmatic Analysis – The general interpretation of language is the focus of pragmatic analysis. It is concerned with leveraging real-world knowledge to derive meaningful interpretations of human language.

### 3.1.4 Feature Extraction Methods

The unstructured text must usually be converted to structured language using a mathematical model or algorithm, as shown in figure 7. To begin, the data must be cleansed to remove any irrelevant characters or words. Formal feature extraction approaches can be used once the data has been cleansed. Feature extraction algorithms are required to convert the text into a matrix (or vector) of features. This transformation task is generally called feature extraction of data. Feature extraction techniques are shown in figure 8. Text feature extraction techniques have two approaches: the traditional approach based on syntactic word representations and the modern approach based on feature learning techniques.

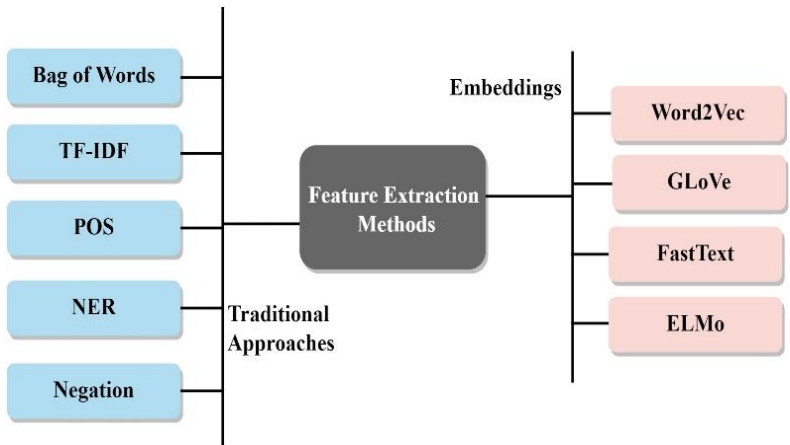

**Fig. 8** Overview of feature extraction methods

*1) Traditional Approach (Syntactic Word Representations) –*
It includes the following methods -
    a) Bag of Words (BoW) — This method is also known as the n-gram method. BOW

counts the frequency of words in the text and generates a scarce vector that represents present words as one and missing words as 0 in the text. Machine learning models use these vectors as input. N-grams are groups of words that appear in close proximity to one another and are merged into a single vector of features. Common representations of N-grams are 1-gram, 2-gram, 3-gram.

b) Term Frequency-Inverse Document Frequency (TF-IDF) – The TF-IDF metric indicates how significant or relevant a term is in a document. It identifies words or sets that frequently appear in a single document yet are uncommon throughout the text corpus. TF-IDF is defined as follows:

TF-IDF = TF * IDF

TF = Term Frequency = no. of term in a document / total terms in the document.

IDF = Inverse Document Frequency = Total No. of documents / no. of documents has term.

So, TF gives the measure of how frequently the term occurs. And IDF gives the measure of how important the term is.

c) Part of Speech Tagging (PoS) – The POS is the word category that showcases its role and helps the NLP work. It includes categories like nouns, verbs, adjectives, and adverbs.

d) Named Entity Recognition (NER) – In the disambiguation, the entities are identified in a sentence, such as the famous person, brand, etc. In the named entity recognition, the entity is identified and categorized as per the date, organization, person, time, location, etc.

2) Modern Approach (Feature Learning Techniques) –

Embeddings - Word embeddings are techniques in which features are learned, and each word from the dictionary is represented to N-dimensional integer. These feature vectors are based on the context of the text dataset. To convert words into intelligible input, many word embedding methods have been developed. Word2Vec, GloVe, FastText, and ELMo are some common deep learning embedding approaches that have been successfully applied.

- *Word2Vec* – (T. Mikolov et al. 2013) introduced the "word to vector" interpretation. It is an upgraded architecture of word embedding. This approach utilizes continuous bag-of-words, skip-gram, and lightweight neural networks consisting of two hidden layers to generate a feature vector for every word. It gives you immediate access to the vectors that represent words. A simple continuous bag of words (CBOW) model looks for words that might appear near one other, whereas skip-gram looks for words that might appear near each other. This approach is a strong method for detecting correlations in a text dataset as well as word resemblance.
- *Glove* – GloVe (Global Vectors for Word Representation) by (Pennington et al. 2014) is another technique for creating word embeddings. It captures a global word-to-word conjunction matrix from a dataset to construct word embeddings. The idea is to establish a context for words i and j in terms of whether or not they occur near N-words apart. The ratio of the co-occurrence probability of two words is encoded in the encoding vector, which is known as the count-based technique. However, this method does not employ the entire corpus. It collects global statistics and improves word representation learning on tasks including word comparison, word similarity, and named entity recognition.
- *Fast Text* - Facebook's AI research team created the FastText library. It is capable of learning and calculating word representations as well as sentence classification. It is a novel technique that solves the morphology issue by allocating a different vector to each word. As a result, each word is represented by an n-gram, which is a collection of characters.
- *ELMo* – Embeddings from Language Models (Peters et al. 2018) is a type of contextual embedding that looks at the words around. It uses a bidirectional LSTM algorithm, which means that the words before and after it are used in both directions. This character-based language model is able to envision the previous or next word, as well as handle words that aren't in the dictionary. It also handles semantical and grammatical characteristics of words and how this differs around language contexts like to develop lexical ambiguity. It advances modern techniques in a broad range of demanding NLP tasks, involving textual context question answering, and sentiment analysis.

## 3.2 Approaches for Emotion Detection–

The most common approaches for identifying text-based emotions are keyword-based, rule-based, machine learning-based, deep learning-based. The approaches for TBED are displayed in figure 3.

*1) Keyword-based approach* – This technique focused on locating keyword occurrences in a given text and comparing them to the annotations registered in the dataset. Figure 9 outlines the method. The initial emotion keyword list is derived from conventional lexical resources WordNet or WordNet-Affect in this method. After that, the dataset is preprocessed. Then, keyword matching is performed between a predefined keyword list and emotion words from the text. After that the power of the emotion keyword is examined. After that, negation is examined to see whether there are any negation cues and what is the scope of, and lastly, the emotion tag calculation is done. In this, we studied (Tao J. 2004, Ma C et al. 2005, Perikos et al. 2013, Shivhare et al. 2015) which are based on a keyword-based approach.

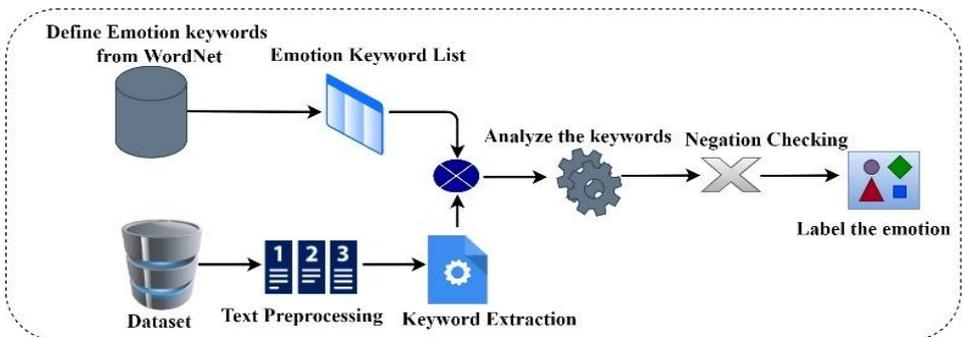

**Fig. 9** Keyword-based approach lay-out

*2) Rule-based approach* – This approach utilizes linguistic rules to recognize the emotions from the text. Figure 10 shows a general overview of the approach. The dataset is first subjected to text preparation. It involves data cleaning, tokenization, POS tagging, etc. Then rules for extracting emotions are built using a statistic, linguistic concepts. Then, probabilistic affinity is stored with each word. Later, the best of the rules has been chosen for the dataset to detect emotion labels. References (Lee et al. 2010, Udochukwu and He 2015, Liu and Cocea 2017) surveyed in rule-based approach.

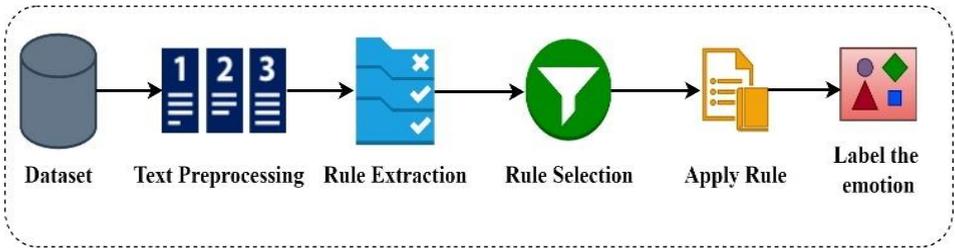

**Fig. 10** Rule-based layout

*3) Machine learning-based Approach* — Approach is depicted in Figure 11. Machine learning-based methods allow systems to learn and improve on their own as a result of their experiences. Text is classified into several emotion classes using machine learning algorithms. There are two types of machine learning algorithms: supervised and unsupervised. Supervised machine learning algorithms are applied in the majority of the studies reviewed. Text preprocessing, such as tokenization, POS tagging, and lemmatization, is usually the first step in this technique. Then the text's useful features are extracted, and only the features having the highest information yield are chosen. The algorithm is then trained using the specified characteristics and emotion labels. finally, the system is utilized to predict emotions from unobserved data using a trained system. The authors examined (Aman et al. 2007, Ghazi et al. 2010, Bruyne et al. 2018, Suhasini et al. 2020, Singh et al. 2019, Allouch et al. 2018, Jonathan et al. 2017) in this approach.

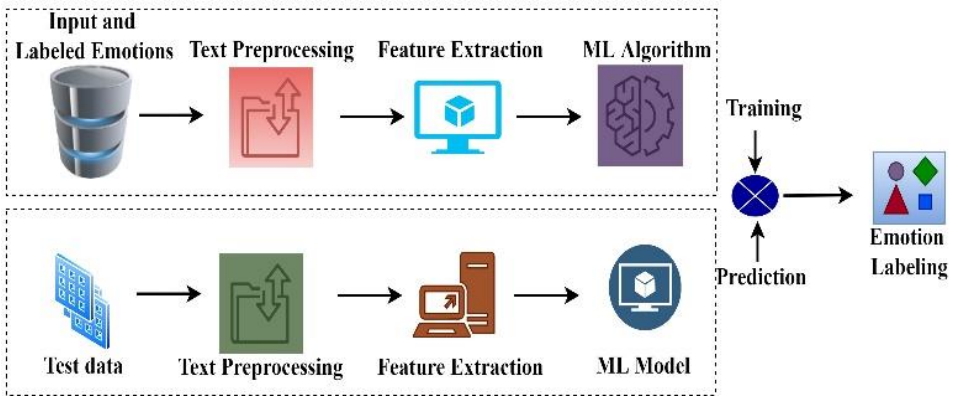

**Fig. 11** Machine learning-based approach layout

*4) Deep learning-based approach* - Deep learning uses neural network layers to learn unsupervised from unlabeled or unstructured data. It is one of the subclasses of machine learning in Artificial Intelligence. Figure 12 depicts the deep learning approach. In the deep learning, neural networks learn complex theories by building from simpler concepts. The emotion dataset is first processed with preprocessing steps such as tokenization, stop words removal, and lemmatization. Then, the embeddings are constructed. In this case, tokens are represented by numbers. Then, using classification, these vectors are provided to deep neural network layers with elements equivalent to emotion labels, where data forms are discovered and used to estimate labels. References (Baziotis et al. 2017, Ezen and Can 2018, Basile et al. 2019, Shrivastava et al. 2019, Xiao 2019, Rathnayaka et al.

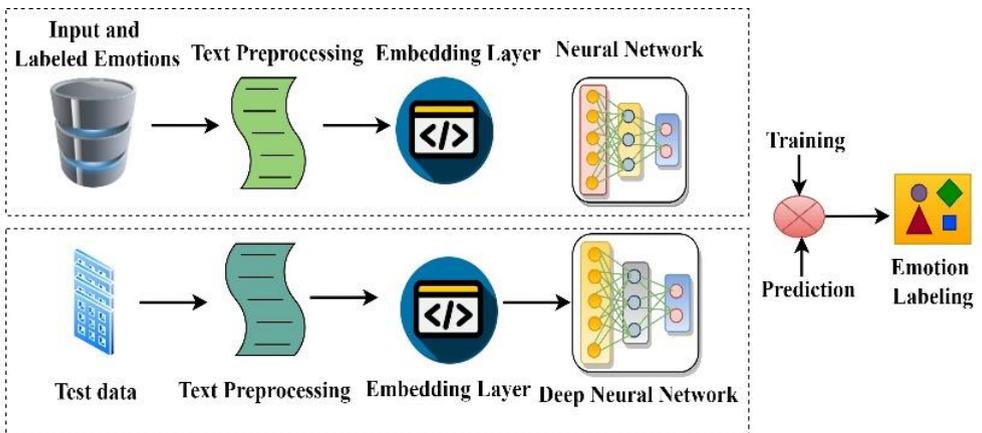

**Fig. 12** Deep learning-based approach layout

2019, Bian et al. 2019, Mohammad et 2021, Jose et al. 2020, Xia and Zhang 2018, Diogo 2021, Taiao et al. 2020, Polignano et al. 2019) surveyed in deep learning-based approach.

### 3.3 Datasets/Corpora

In this section, the authors discuss widely used datasets for TBED, which is the primary focus of this article. In addition, this section informs researchers on recent advancements in the domain of TBED. In TBED, researchers can either create their own datasets or use publicly available datasets. In this section, the authors present some publicly available and useful datasets with a reliable labeling or annotating process. Figure 13 gives a representation of different datasets with application domains used. These datasets are widely used by researchers in TBED and are described as follows:
*1) ISEAR (The International Survey on Emotion Antecedents and Reactions):*
This dataset was created in 1997 by (Scherer and Wallbott 1994). It's one of the first

curated datasets using emotion labels, and now it's free to download. The information was gathered by a group of psychologists who worked on the ISEAR project in the 1990s. There are seven major emotions included in this dataset: joy, anger, fear, disgust, guilt, sadness, and shame. 3,000 students from 37 countries, including psychologists and non-psychologists, were asked to explain situations in the form of emotions they had encountered in this study.

*2) Cecilia Ovesdotter Alms' Affect data:*
(Alm and Sproat 2008) developed this dataset in 2005. It includes 185 stories by Beatrix Potter, Hans Christian Andersen, and the Brothers Grimm. It has 15000 sentences labeled with the emotions fear, anger, disgust, sadness, happiness, and surprise. It also has positively surprised, negatively surprised, or neutral if no emotion is displayed.

*3) SemEval (Semantic Evaluations):*
The database Semantic Evaluations (SemEval) (Rosenthal et al. 2019) contains Arabic and English data. According to the data, this dataset has three versions. SemEval 2007 is a collection of news headlines from various publications, including BBC News, The New York Times, CNN, Google News, and others. Surprise, anger, fear, sadness, disgust, joy are all annotated with one or more emotions in each headline. SemEval 2018 is made up entirely of tweets. Anger, fear, disgust, love, joy, optimism, sadness, pessimism, trust, and surprise are among the eleven emotions expressed in each tweet. Distinct test, trial, and training datasets are provided for Spanish, Arabic, English, tweets. SemEval 2019 is made up of textual conversations between 2 persons. Each interaction is categorized into one of four groups: anger, sadness, joy, or others. The third turn of the discussion is used to categorize emotion labels. Distinct training, trial, and test datasets are available.

*4) Crowdflower:*
This dataset (https://data.world/crowdflower/sentiment--in-analysistext) contains tweets and their labels for the emotional content of texts. It was developed in 2016. It has a substantial number of labels across 13 labels. It includes emotion labels: boredom, anger, empty, fun, enthusiasm, happiness, love, hate, neutral, sadness, relief, worry, surprise. This dataset has 40000 tweets.

*5) Emobank:*
Over 10,000 words have been dimensionally labeled using the Valence-Arousal-Dominance (VAD) emotion representation paradigm. (Buechel et al. 2017) developed it in 2017. These sentences were collected from a wide range of sources, including essays, news headlines, blogs, fiction, newspapers, travel guides, and letters written by writers and readers. Ekman's basic emotion model has been used to categorize a subset of the dataset, making it eligible for dual representational techniques.

*6) Aman:*
This corpus is made up of blogs that were obtained using Ekman's six fundamental emotions as clue words, e.g., In the pleasure category, the words happy, delighted, and enjoy were chosen. It was created by (Aman et al. 2007) in 2007. The sentence-level annotation was done, and there are eight different emotion classes to choose from, including two new ones: mixed emotion and no emotion. It has 1466 sentences labeled with mixed emotion and no emotion, sadness, surprise, happiness, fear, disgust, and anger. Sentences depicting neutral feelings are defined by the word "no emotion."

*7) The Valence and Arousal Facebook Posts by Preotiuc-Pietro:*
(Preotiuc and Pietro 2016) created this data set in 2016. It has been labeled on two scales that are independent of one another: The polarity of the affective content of a post is represented by valence (or sentiment), which is assessed on a nine-point scale from 1 (very negative) to 5 (neutral/objective) to 9 (extremely positive) (very positive). Arousal (or intensity) is a nine-point scale that ranks the strength of affective content from 1 (neutral perspective) to 9 (very intense) (very high). Facebook status make up of this corpus. There were 3120 posts in the original data set.

*8) MELD (Multimodal Multi-Party Dataset for Emotion Recognition in Conversation):*
It is a multimodal corpus that includes data from textual, audio, video data. It has around 1400 conversations and 13000 statements from the TV series Friends, which are labeled as anger, grief, disgust, surprise, fear, joy, and neutral in dialogues. It was developed by (Poria S et al. 2018) in 2018.

*9) CBET (the Cleaned Balanced Emotional Tweets):*
In 2015, (Gholipour Shahraki 2015) compiled this dataset from Twitter. He used hashtags to detect tweets with one of nine emotions: joy, anger, sadness, love, disgust, surprise, fear, guilt, and gratitude. The CBET is divided into two sections. One section has only one labeled tweet, denoted as single labeled tweets. This section has 76,860 tweets, with 8,540 for each emotion, and is completely balanced over labels. Double-labeled tweets express two emotions in the text, which are found in the smaller section. This part has 4,303 tweets and is unbalanced since not all combinations of emotions occur equally and frequently.

*10) GoEmotions:*
This dataset comprises 58K Reddit comments from 2005 (the start of Reddit) to January 2019. In addition, this dataset has labeled for one or more of 27 emotion(s) or Neutral as follows: admiration, amusement, approval, annoyance, anger, curiosity, confusion, caring, desire, disgust, disapproval, disappointment, embarrassment, excitement, fear, joy, grief, gratitude, love, sadness, nervousness, optimism, pride, remorse, relief, realization, surprise. (Dorottya Demszky et al. 2020) developed this corpus in 2020.

*11) Amazon:*
The work of (Blitzer et al. 2007) in 2007 resulted in the name "Amazon." It's a product review dataset compiled from the Amazon website. It features 2,000 reviews in four distinct sectors, including DVDs, books, electronics, and kitchen products. This dataset is balanced because each section has equal positive and negative reviews. Each sample in the dataset contains specific information about a review, such as a title, the rating ranging from 0 to 5, and date of the review, and the substance of the review.

*12) IMDB (Internet Movie Database):*
It's the unlabeled version of the polarity dataset. The IMDB dataset (https://datasets.imdbws.com/) was created to aid in the classification of binary sentiment in movie reviews. There are equal positive and negative reviews on IMDB. It is split uniformly in the training and test datasets, with each receiving 25,000 reviews. It was created in the year 2011. Its data consists of 27,886 unprocessed and unlabeled HTML files.

*13) The Movie Review (MR) dataset:*
It is a dataset of movie reviews curated to detect the sentiment related to detailed reviews and to determine positive or negative reviews. It contains 10,662 review sentences with the equivalent count of positive and negative instances. This dataset was developed in the early 2000s by (Bo Pang and Lillian Lee 2002).

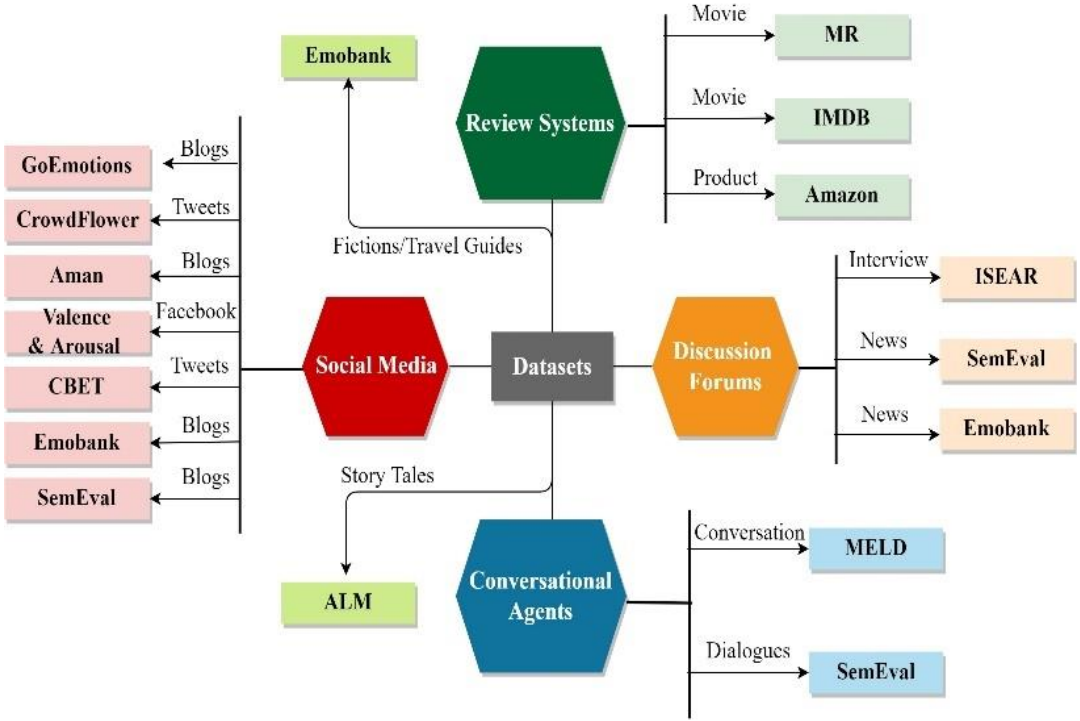

**Fig. 13** Datasets with topics and domains

## 3.4 Application Domains

Text-based emotion detection has wide applicability to various application domains. This wider range of application domains includes management and marketing, user interaction, healthcare, education, finance, public monitoring, etc. An abundant amount of text data is generated and analyzed to make decisions in many disciplines such as online social media, customer or product review systems, recommendation systems, discussion forums, and conversational agents. We are focusing on online social media, conversational agents, and review systems.

### 3.4.1 Conversational Agents

Conversational agents are referred to by various terms, including chatbots, dialogue systems, intelligent virtual assistants, intelligent virtual agents, smart bots, interactive agents, digital assistants, and relational agents. A chatbot is a human-computer interaction model with an artificial intelligence program (Eleni et al. 2020). A chatbot is described as a "computer software meant to simulate interaction with human users, particularly over the Internet, by using spoken or written natural language as a primary means of communication.". Conversational agents, often known as chatbots, mimic human interaction and are utilized in various areas such as business, education, healthcare, entertainment, and many more. The primary goals of chatbots are entertainment, social contact, and novelty interaction, with a strong emphasis on productivity. In chatbots, there has been a tremendous requirement to simulate human-like characteristics and behavior. Chatbots are usually depicted, built, and developed as a flow of communication between multiple components as Natural Language Understanding (NLU), Dialogue Management (DM), and Natural Language Generation (NLG). The NLU generates a semantic representation of a user text, such as intent or purpose, to extract the "meaning" of the text. The NLU's main role is parsing, which involves taking a collection of words, generating a linguistic formation for the words, and storing in the knowledge repository. The second component is dialogue management, which formalizes communication structure and performs contextual understanding by assessing word disambiguation and identifying phrases connected by references and supports domain knowledge supervision. It can match expressions (emotion) using lexicon and handling synonyms, negation, and intensity based on the presence of terms (emotional). The third component is NLG or response module. Commonly, Chatbot responses are restricted to providing a direct response to the participant's question. NLG gets a communication response from the dialogue manager and creates a textual depiction that corresponds. Language generation interprets gist by selecting the syntactical form and words required to represent it.

In business, conversational agents provide availability, scalability, and cost-effectiveness by enhancing market competitiveness and service quality (Adikari et al. 2019). These chatbots also improve customer engagement by offering friendliness, comfortness, flexibility, and efficient assistance. Chatbots should provide customers with more engaging responses, directly addressing their issues. Users perceive chatbots as companions rather than simple assistants. The majority of user requests are emotional rather than informative. Chatbots gained the capacity to respond emotionally to customers due to machine learning and sentiment analysis evolution. (Adikari et al. 2019) discussed an artificial intelligence-based cognitive model to detect emotions in industrial chatbots using word embedding, Markov chains, and NLP. Conversational agents are increasingly being used in industrial applications, from simple conversation interfaces to sophisticated human aides. Industries are infusing human-like qualities, features, and behaviors into conversational agents to increase productivity. A chatbot implemented by (Thomas 2016) responds to the users with AIML, and LSA uses a dataset of Frequently Asked Questions. (Xue et al. 2018) designed a neural-based conversational agent that uses Bi-LSTM with attention to reply to repetitive questions. (Sun and Zhang 2018) developed a personalized recommendation system. It includes three key components: a deep belief network as a tracker, an LSTM network, a deep policy network. In order to optimize user experience, it also focuses on the personalization of the recommendation system. They used Factorization Machines (FM) that glues the strengths of Support Vector Machines (SVM) and factorization models.

In the Healthcare field, conversational assistants are used for diet management, medication adherence, mental wellbeing, and physical activity (Fadhil et al. 2019). Furthermore, due to the low learning curve, users, particularly the elderly, are comfortable with chatbots. In (Fadhil et al. 2019) developed a CoachAI, a health coaching application driven by a conversational agent with a supervised machine-learning model. In the healthcare field, conversational agents could assist users by presenting appropriate information about an illness or explaining the findings of clinical tests (Joao et al. 2019). Clinicians use conversational agents to help them identify symptoms and improve their assessment abilities, diagnosis, interview procedures, and interpersonal communication. (Pacheco et al. 2021) reviewed smart conversational agent systems to detect neuropsychiatric disorders. Affect and cognition are the human abilities most desirable by chatbots to accomplish change behavior in humans (Pereira et al. 2019).

In the education field, educational conversational agents are designed to facilitate and support online learning. These chatbots enable interactions between the learner and the content; thus, the agent's role is to deliver the instructional content (Song et al. 2017). Chatbots systems are employed for literacy tutoring chatbots, security training chatbots. Chatbot assists students (Hien et al. 2018) in administrative concerns for example applying for a course, finding out about examination schedules, results, and other academic particulars. In (Jia and Jiyou 2004), a conversational agent was developed to converse with English learners using a basic method of logical reasoning and inferences based on semantic and syntactic analysis.

Chatbots can play different roles, such as recommenders, information providers, tutors, advisors, conversational partners, entertainers, and can be used for multi-party interaction, decision-making, conflict, and opinion resolution. Pattern matching and machine learning are two approaches used to construct a chatbot, based on the algorithms and methodologies

*1) Pattern matching* – It is also called a rule-based approach. Pattern matching algorithms compare user input to a rule or pattern and choose from a set of replies a preset answer. The context can also influence which rules are used and how the response is formatted. The first chatbots to use the pattern matching approach were ALICE and ELIZA. It uses logic programming clauses and if-else patterns for applying to reason. It traverses the tree structure made up of rules to find the next action to perform.

*2) Machine learning* - For chatbot systems, various machine learning approaches exist, each with its own set of learning scenarios. Each learning scenario produces a conversational action to be taken next. Machine learning algorithms have grown in popularity since they can choose which conversational act to use at any time, based on prior knowledge removing the need for domain experts to introduce policies and rules to the system regularly. Bayesian networks, neural networks, Markov models are different mechanisms used in the machine learning approach. For example, the development of chatbots based on Artificial Neural Networks (ANNs) uses score assignment and chooses the most expected response from a collection of responses in retrieval-based models.

While in Generative models, on the contrary, the response is synthesized using deep learning approaches.

Chatbots have several disadvantages and risks in addition to their tremendous benefits. Data security is an important issue in conversational agents for both user and agent service providers. Chatbots often fail to detect the intentions of their speaker. Failure to detect the intent of the user may create frustration for the user. And its adverse effect may prove ruinous to the service provider. Long responses covering the important information that may discourage users and lead to a break in the conversation. Intention classification can also be affected by user errors in spelling. Misused phrases, subtle sarcasm, language impairments, usage of slang, and syntax faults are all examples of failures caused by user input. The vast majority of chatbots are focused on business, notably e-commerce; other areas, such as healthcare, finance, management, education, have received less attention.

**3.4.2 Online social media –**

The majority of the world's population uses popular social networking sites or applications such as Facebook, Instagram, Reddit, Twitter, YouTube, and WhatsApp. Furthermore, microblogging services such as Instagram, Reddit, and Twitter are immensely popular. Consequently, we may state that postings that include text, images, audio, and video, online media has developed as a strong medium for expressing, communicating, and sharing people's thoughts, opinions, perspectives, views, and including native to worldwide issues and topics. So, it becomes important to analyze the emotions behind people's opinions, thoughts, views, and perspectives. This study will be useful in finding the impact of OSM on different online trends(positive/negative/influential/biased), social behavior, and individual social conduct. Feelings, emotions, character traits, and their impact on societal trends can all be found using TBED. There are various online social media application areas such as business intelligence, behavioral analysis, epidemics, event detection, crime detection, sentiment, opinion, and emotion analysis, recommendations.

The authors in (Sailunaz et al. 2019) developed a recommender system based using sentiment analysis and emotion detection employing machine learning algorithms on the Twitter posts. By using the Naïve Bayes algorithm, anger, disgust, joy, fear, surprise, sadness, and neutral emotions are detected by their system. While positive, negative, and neutral polarities are identified in sentiment analysis for each Twitter post. Their system calculated users' influence scores using a variety of user- and tweet-based factors. And used this data to create generalized and individualized suggestions for users based on their Twitter behavior. (Pool et al. 2016) explored emotion classification using Facebook reactions in a distant supervised context. The authors used Support Vector Machine (SVM) with a simple BoW to classify the emotions like anger, joy, sadness, surprise. (Douiji et al. 2016) proposed a system coping with the complications of writing style in chats and language progress to detect users' emotions from their textual interactions from YouTube comments. Based on a dataset produced from YouTube comments, their approach used an unsupervised machine learning technique to do emotion classification. The authors utilized Latent Semantic Analysis, and its Probabilistic version, Non-negative Matrix Factorization to detect sadness, joy, fear, surprise, disgust, Anger from YouTube comments.

In (Kumari et al. 2021) authors discovered a text-based cyberbullying detection system in social media using semantic autoencoders. They proposed a method named semantic-enhanced marginalized stacked denoising auto-encoder (smSDA) via semantic dropout noise based on word embeddings of domain knowledge and sparsity constraints. They evaluated the performance of the system using Twitter and myspace corpora of cyberbullying. In (Wang et al. 2017) authors developed a model to discover a group of closely connected people and highly involved in their sentiments about a product/service. They introduced the concept of sentiment community using the optimization models of semi-definite programming (SDP). The authors developed an influential activity maximization model to identify the influential node. Independent cascade and linear threshold, these diffusion models are used in their activity maximization model. Authors in (Song et al. 2017) studied the spam messages on social media sites. They proposed a novel methodology for social spam detection using labeled latent Dirichlet allocation (L-LDA) on the word-based, topic-based, user-based features, and accuracy improved using incremental learning. In (Gupta et al. 2021) the authors present a research approach that effectively measures population opinions and identifies misunderstandings, informing public health communications in epidemics. They studied the ability of weather to affect the spread of COVID-19 pandemics. Their work was based on Twitter data and used Gradient Descent Support Vector Machine (GDSVM), random forest, naïve Bayes, and TF-IDF machine learning algorithms. Topic modeling is used to find out major topics of discussion in the pandemic. Authors in (Sun et al. 2015) designed a social regulation method that incorporates social network information to benefit recommendation systems. The authors considered the social relationship among the users for improving the accuracy of the recommendation system. They used a bi-clustering algorithm to calculate the trusted friends in the network.

Many difficulties must be overcome while identifying emotions via social media. First, the casual writing style of users in social media is a big challenge. Grammatical and spelling mistakes, as well as slang language, sarcasm, and irony, may be found in social media posts. Simultaneously, the use of informal language and short messages has been studied, but only to a limited extent. Second, the expression of human emotions and the texts that convey them are subjective and ambiguous. As a result, correctly inferring and interpreting the author's emotional states is challenging. Third, emotions are multifaceted constructs with unclear confines and varying expressions. As a result, human mapping is important. Finally, affective expression is difficult for automated systems to understand. Because of the wide range of topics discussed on social media, manually creating a complete dataset of labeled data that includes all conceivable emotional circumstances is difficult.

### 3.4.3 Review systems

BI (Business Intelligence) (Sandra et al. 2021) is a set of methodologies, frameworks, and approaches for transforming data into actionable information that helps businesses run more efficiently. It is, in fact, a set of apps and services that convert data into usable knowledge and experience. The use of the BI paradigm to social media data benefits businesses of various domains. Social media platform data can benefit various business domains. Business-related information gathered from product/customer reviews, tweets, likes, shares, comments of specific and related business data and disseminated through online social media platforms can assist businesses in producing high-quality products and services. Especially in product design and development, early feedback, detection, and conceptual concepts from customer study are crucial. A good user experience with the product, system, and service is gradually recognized as a significant product/service design advantage. In general, user experience/feedback (UX) refers to how a user feels about utilizing a product, system, service, or object (Bai et al. 2019). Customer holding and trustworthiness are critical components of business success. Because there is widespread agreement that customer satisfaction is strongly linked to slow customer destruction, many business strategies focus on analyzing and improving customer satisfaction. According to the application area, emotion recognition from product and customer reviews has been researched more than service reviews.

The authors of (Bai et. al 2019) examined how user input data may be retrieved and restored from online customer reviews to help product design. A multi-aspect proposed framework is developed as an operational technique for highlighting the key components of user feedback in product design. The authors also suggested a technique for constructing a knowledge base of user comments from online consumer reviews. User feedback discovery, which extracts user feedback data from a single review, integrates with similar group data, and user feedback network formalization, which establishes causal linkages among user feedback data groups, are the three stages of their approach. By analyzing a large number of online customer data, this study analyses the feasibility of automatically uncovering relevant user feedback data and their correlations for product creation and strategic business planning. The overall goal of (Monireh et al. 2018) research is to create an automatic tool that analyses customer reviews of a product and determines how they feel about various characteristics of that product. Authors offer the Pros/Cons Sentiment Analyzer (PCSA) framework for collecting sentiment knowledge from pros/cons reviews using dependency relations. Feature extraction using syntactic rules and determining the opinion's polarity according to their emotional power from pros/cons reviews highlight their work. Authors in (Cambria et al. 2020) used proposed domain adaptation using genetic programming in product reviews. The authors proposed a framework, Genetic Opinion Adaptation Learning (GOAL) which combines convolutional deep belief network (CBDN) with Genetic Programming (GP). Authors use a variable-length tree called a genetic program to model each text's features. At internal nodes in the tree, mathematical operators such as '+' or '-' can denote the polarity of phrases. The proposed model is tested using Amazon product reviews for a variety of products and languages.

Even while this field's research isn't inextricably connected to the crucial user feedback idea "emotion" (e.g., happiness, joy, fear, surprise, and disgust), user sentiment like a neutral, negative, or positive does refer to feelings, albeit not to the same extent. Therefore, it is needed to present a possible map from sentiments to feelings by further leveraging the language elements in the feedback/ evaluations, such as sorting them based on emotion categories. Emotion detection from review text has mainly been done in movie review systems, hotel review and recommendation systems, E-commerce review systems, agriculture product review systems.

## 3.5 Classifiers

After generating feature vectors and creating word embeddings next significant step is to classify the emotions. Deep learning and machine learning are the two classification techniques used by most researchers. Overview of different classifiers used in text-based emotion detection is shown in figure 14. This section describes different deep learning and machine learning classifiers used to classify emotion purposes in detail.

### 3.5.1 Machine Learning Classifiers

Machine learning classifiers are used extensively and comprehensively in the field of text-based classification. These classifiers work on annotated/labeled datasets, so known as data-driven (Nantheera 2020). Computing methods in machine learning are used to directly learn knowledge from large amounts of training examples/samples. These algorithms adaptively arrive at optimal solutions and generally enhance their effectiveness as the number of samples supplied for learning increases. There are three different categories of learning classifiers: supervised, unsupervised, and reinforcement learning. A supervised learning classifier contains inputs and desired output labels, whereas an unsupervised learning classifier deals with unlabeled data. Reinforcement learning classifiers learn from trial and error and are self-supervised means a limited set of annotated data provided for labeling and usually a large amount of unlabeled data. Decision Tree (DT), K Nearest Neighbor (KNN), Support Vector Machine (SVM), Random Forest (RF), Naive Bayes (NB), Linear Regression (LR), Multinomial Naive Bayes (MNB) are some most preferred machine learning classifiers.

*Decision Trees (DT):*
A decision tree is a rule-based supervised classifier (Allouch et al. 2018). It is a tree structure made up of nodes and branches/directed edges. Each node signifies an attribute/feature, and each branch signifies the decision rules or tests on attributes (Hasan et al. 2019), and each leaf node demonstrates a class label/output. There are three steps for decision tree learning (Adivi et al. 2019): features choosing, decision tree generation, and pruning. Simple decision rules from the features are used for the learning process (Shayaa 2018). The decision tree is a frequently used classifier for regression and classification. This method represents decision-making through a tree-like structure (Pradhan et al. 2016). When using a DT, figuring out which attributes to be utilized, what criteria to be utilized for splitting, and when to terminate are crucial factors. Throughout this procedure, all attributes are examined, and several division points are explored and tried. The split that costs the least is picked to determine the optimum path. Pruning a tree can boost its performance even further. Pruning is the process of deleting less important elements from a tree to reduce its complexity and, as a result, increase the predictive power by reducing overfitting. (Kaur 2012) utilized the decision tree to investigate human emotion variance (Kaur 2012). The decision tree can be used to examine and classify many types of emotions. To detect the emotions of children with diverse disabilities, Kaur employed outlier analysis.

*Random Forest (RF):*
Random Forest is a supervised classification algorithm. It is built using the decision tree approach. The term "forest" refers to a collection of decision trees that have been combined to produce a more accurate, robust prediction and enhance the overall result. Because of its simplicity, this approach is flexible and straightforward to use, so it is used for classification and regression problems. The trees continue to grow, and random forest adds randomness by observing the finest attributes within a subset of random features rather than only investigating the most significant ones. It helps increase variation in this way, leading to improved models and results (Fang and Zhan 2015). Boosting and bagging are the two most used classification approaches (Segnini et al. ). Bootstrap aggregation is referred to as "bagging." It combines many learners/predictors to minimize the decision tree's variance. Random forest trains N decision trees on diverse random subclasses of the data and calculates the average of all the predictions for final prediction (Abu et al. 2018). Boosting method is used to create a strong classifier from a set of weak classifiers. Boosting try to deal with errors from the prior decision trees by forming a group of sequential predictors. Boosting techniques can be used to track down the model that failed to provide an accurate prediction. Overfitting is less of an issue with boosting approaches. Ensemble learning generates predictions by averaging the decisions of numerous predictors. It produces better predictions compared to a single decision tree.

*k-NN (k - Nearest Neighbor) –*

k-NN is one of the simplest categories. The nearest neighbor of K is the meaning of k nearest neighbor. The central impression of the k-NN algorithm is that a sample is classified based on the majority votes of its neighbor, with the sample classified to the class that is most common between its k nearest neighbors. The method determines the type of sample classified based on one or more recent samples (Kaur and Duhan 2015). It used Euclidean distance to determine the distance of an attribute from its neighbors (Wu et al. 2008). The KNN algorithm is a slow, non-parametric algorithm that makes no assumptions about the underlying data distribution and does not require any explicit training before classification (Shayaa 2018). It is used for classification and regression.

*Support Vector Machine (SVM) –*

SVM is a supervised learning model. It is used to solve problems involving classification and regression. It's mostly used to organize linearly separable data, although it can also handle non-linear data in a high-dimensional feature space. It constructs a set of hyperplanes (decision boundaries) in high-dimensional space with the largest margin to classify the data (Kaur and Duhan 2016). It aims to improve the classification's robustness by using the largest margin. In most classification problems, more complex structures are required to achieve optimal separations. A nonlinear separation is required to better identify the clusters in most circumstances than a linear line. SVM overcomes this challenge by employing kernel functions, which are mathematical functions, to reorganize the data set. To do the linear separation, the technique entails plotting the input data in a new space and translating it into a higher dimensional feature space. SVM has the advantages of being effective in high-dimensional space, memory efficiency, and versatile as various kernel functions available (Zahid et al. 2020). However, overfitting may arise if the count of features exceeds the count of samples. To categorize reviews, (Cheng and Tseng 2011) employed two multiclass SVM-based techniques: To assess the reviews' quality, researchers used Single-Machine Multiclass SVM and One-versus-All SVM (Cheng and Tseng 2011).

*Naïve Bayes (NB)-*
The Naïve Bayes classifier is one of the simplest classifiers used for text classification. It is derived from the Bayesian theorem of probability, and the Naïve Bayes assumption assumes that the features of an object are conditionally independent, given the class label of the object. There are many variations, such as the Multinomial Naïve Bayes, the Gaussian Naïve Bayes, and the Bernoulli Naïve Bayes, all of which operate on the probability of the occurrence of words in the text. This classification is straightforward and effective, and it's especially well-suited to inputs with a lot of dimensions. Naive Bayes outperforms more advanced categorization systems despite their simplicity (Sharupaet 2020). The Bayes rule, for example, determines the likelihood of features appearing in each class in Bayesian analysis and then delivers the most probable class. This approach was previously described in (Tang et al. 2009). They also introduced a Multinomial Naive Bayes (MNB) method for resolving existing uncountable and meaningless features, which takes into account the amount of positive and negative words while computing weights and eliminates irrelevant words while selecting features.

### 3.5.2 Deep Learning Classifiers

Deep learning classifiers come under unsupervised learning. However, it can be supervised or semi-supervised too. Deep learning classifiers learn and extract features automatically, resulting in enhanced accuracy and performance. Convolutional Neural Network, Recurrent Neural Network, BERT, Bi-LSTM, GRU, and pre-trained models are some most preferred deep learning classifiers.

*Convolutional Neural Network (CNN) –*
CNN is widely seen in Recommender Systems, Computer Vision, and NLP applications. It is a convolutional and pooling or subsampling layer of a deep feed-forward neural network that feeds data into a fully connected neural network layer (Zhang et al. 2018). Convolution layers obtain features by filtering input data, and several filters are merged to get outputs. With pooling or subsampling, layers feature resolution is lowered to improve the CNN's distortion and noise resistance. Minimizing the dimension of the result from one layer to the next subsequent layer is called pooling. Various pooling algorithms are utilized to lower the number of results or outputs while maintaining essential attributes. The commonly used pooling approach is max pooling, which selects the largest factor in the pooling window. Fully connected layers carry out classification tasks. When it comes to identifying local and position-invariant patterns, CNNs perform well. CNN's have been employed efficiently in text classification (Shervin et al. 2021).

*Recurrent Neural Network (RNN) –*
Researchers used a recurrent neural network for text mining and classification (Sutskever et al. 2011). It is a kind of neural network architecture in which previous data points in a sequence are given higher weight by RNN. As a result, this strategy effectively classifies text, string, and sequence data. Furthermore, in a very sophisticated approach, an RNN examines the information of prior nodes, allowing for a greater semantic understanding structure of a dataset. RNN (Britz 2015) consists of feedback loops in which the connections between neurons create a directed cycle. The basic role of RNN is to process sequential input using internal memory obtained during directed cycles. RNN can recollect and apply prior knowledge to the next element in the input sequence, unlike traditional neural networks. When gradient descent error is backpropagated across the network, RNN is prone to vanishing gradient and exploding gradient difficulties. As a result, RNN primarily use LSTM or GRU for text categorization.

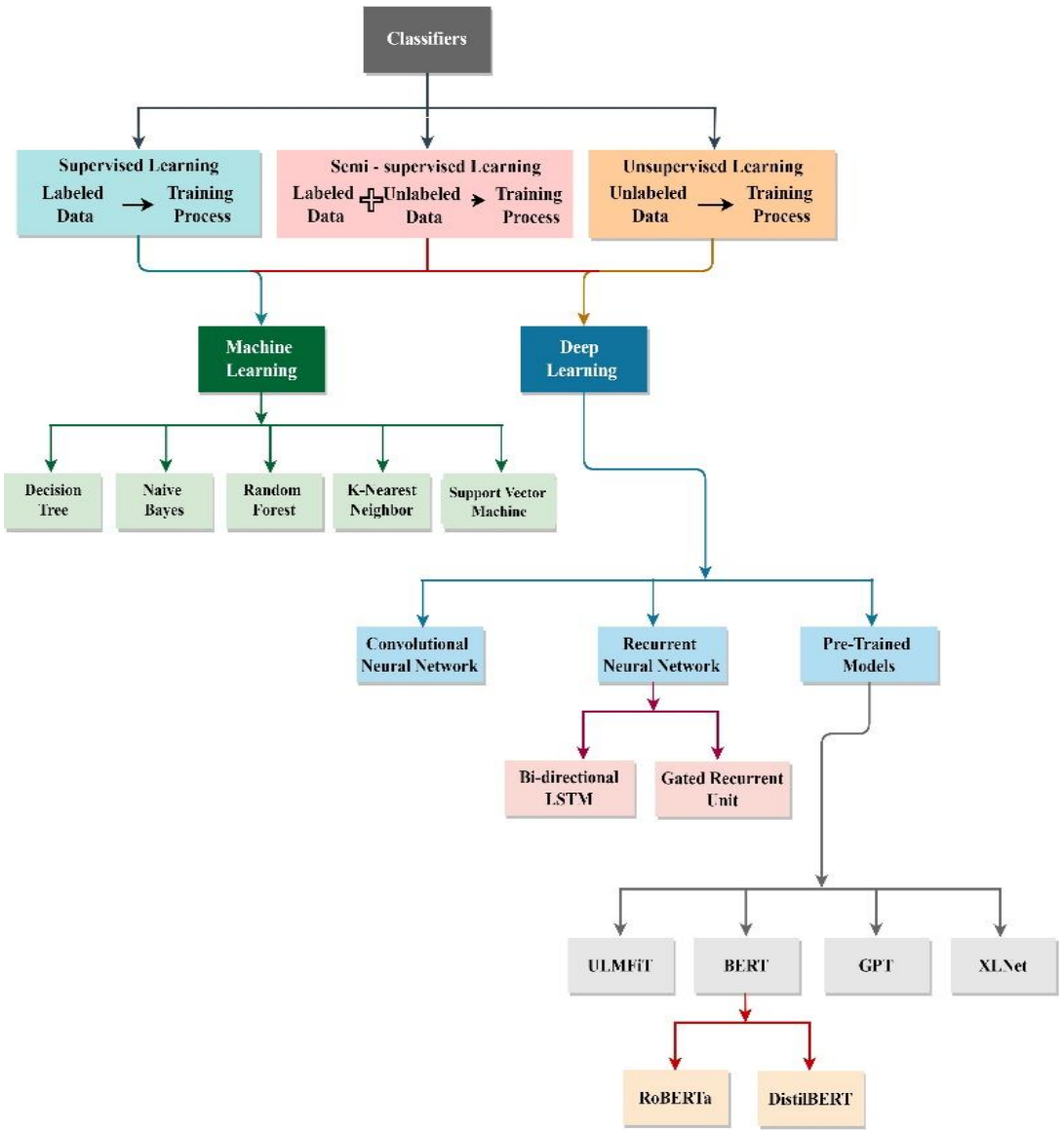

**Fig. 14** Overview of Classifiers

*LSTM -*
S. Hochreiter and J. Schmidhuber (Hochreiter et al. 1996) invented LSTM, which has since been expanded by a number of research experts (Graves et al. 2005). The LSTM is a form of RNN that solves issues more successfully than the basic RNN by retaining long-term dependency. When it comes to addressing the vanishing gradient problem, LSTM proves to be useful (Pascanu et al. 2013). LSTM also had chain-like architecture like RNN, but it uses many gates to closely limit the extent of information permitted into each node state. The input is preprocessed to rearrange it for the embedding matrix. The next layer is the LSTM, which has 200 cells. The final layer comprises 128 text classification cells and is fully connected. The final layer uses the sigmoid activation function to reduce the 128-height vector to a one-height output vector if there are two classes (negative, positive) to forecast. Bidirectional LSTM networks are one sort of LSTM network. Bi-LSTMs are made up of two hidden layers. The input sequence is processed forward using the first hidden layer and backward using the second hidden layer. The output layer combines the hidden layers, allowing it to access the past and future context of each point in the sequence. The LSTM and its bidirectional variations have proven to be extremely useful. They may learn when and how to forget particular pieces of knowledge, as well as when and why not to use specific gateways in their architecture. A Bi LSTM network has the advantages of a faster learning rate and greater performance (Schuster et al. 1997).

*GRU –*
(J. Chung et al. 2014) and (Farzi et al. 2016) proposed GRUs as a gating mechanism for RNN. The GRUs architecture is a simpler version of the LSTM. A GRU, on the other hand, differs from an LSTM in that it has two gates and no internal memory. In addition, a second non-linearity is not used. (Cho et al. 2014) presented the Gated Recurrent Unit, a recurrent neural network that attempted to solve the vanishing gradient problem (GRU). Because both are built similarly, GRU is a variant of the LSTM. To overcome the vanishing gradient problem, GRU employs an update and reset strategy. The model uses the update gate to decide the amount of previous data from prior time steps that should be utilized in the next future steps. The reset gate decides the amount of data is discarded. A Bi-GRU network that is bi-directional is a variant of GRU (Bi-GRU).

*Transformer Models –*
Transformer-based deep learning models use the self-attention process, weighing the significance of each portion of the input data differently. Comparison analysis of different pre-trained models is shown in table 6.

*BERT* – The BERT stands for Bidirectional Encoder Representations from Transformers. It is a language transformation methodology launched by Google (Devlin et al. 2019). It is "deeply bidirectional". It acquires the extensive interpretations of texts by taking into consideration right and left contexts equally. This method is utilized to train general-purpose language models on big datasets and solve NLP problems. The pre-training and fine-tuning steps are involved in employing BERT. BERT model is trained on unlabeled data in the pre-training phase. The model is then fine-tuned for specific NLP tasks after being initialized with the pre-trained parameters. The cost of fine-tuning the BERT model is substantially lower in terms of system resources. BERT employs the same architecture for a variety of activities. The Transformers (Vaswani et al. 2017) are used to construct BERT. The model is available in two sizes: BERT-base and BERT-large. BERT-base has

12 Transformer blocks with a hidden size of 768 and 12 self-attention heads, whereas BERT-large has 24 Transformer blocks with a hidden size of 1024 and 16 self-attention heads. The English Wikipedia and BooksCorpus datasets were used to train BERT. DistilBERT focuses on shrinking the dimensions of the BERT while maintaining performance (Sanh et al. 2020), to enhance performance, RoBERTa (Liu et al. 2019) exploits fully, hyper-parameters of BERT and XLNet acquires contexts in both directions using all permutations of the factorization order (Yang et al. 2020), are three BERT variants that have recently been introduced.

*ULMFiT* – ULMFiT (Howard et al. 2018) was developed at the beginning of 2018. ULMFiT was the first "pure" transfer learning application. It employs AWD-LSTMs, an LSTM version that employs Drop Connect for improved regularization and improves using averaged stochastic gradient descent (ASGD). This model has 400 dimensions in the embedding layer and has three LSTM layers, with 1150 hidden units in each of them. On top of this approach, the authors stacked a SoftMax classifier of size 50 to pre-train the model. During fine-tuning, this last layer is supplemented with a task-specific layer. The vocabulary is constrained to 30k words. The model was evaluated on several data sets IMDb, TREC-6, Yelp-bi, AG's news, Yelp-full, DBpedia.

*GPT* – The Open AI GPT (Radford et al. 2018) design is purely attention-based, with no recurrent layers. By combining learned position embeddings with byte-pair encoded token embeddings and incorporating these embeddings into a cross-layer transformer decoder framework with the purpose of standard language modeling, pre-training is achieved. The model performs at each step by utilizing a decoder architecture. It has access to the tokens that came before it in the sequence. As a result, the GPT model is a one-way attention model. GPT uses transformer decoders to model language using a semi-supervised learning technique. The GPT, primarily used for text representation, is formed by the transformer decoder with 12 attention heads and 12 transformer layers. Pre-processing enormous unlabeled datasets, such as the BooksCorpus dataset on the restricted supervised datasets, trains and fine-tunes them. In 2019, the Open AI team released GPT2 (Radford et.al 2019), a scaled-up version of GPT.

*XLNet* - In XLNet (Yang et al. 2019), Transformer-XL and Permutation Language (PLM) models are employed. XLNet is a language model that is both auto-regressive and auto-encoding. BERT masks the data and uses a bi-directional context to try to predict the masked data, whereas XLNet uses the permutation objective. In a sentence, complete distinct combinations of words are trained, as the model may learn bi-directional context with the help of PLM. The techniques of positional encoding and recurrence are then applied. For maintaining the location information of the token that is to be predicted, it employs two-stream self-attention. XLNet is pretrained and fine-tuned on English Wikipedia, Common Crawl, Clueweb 2012-13, Giga 5, and BooksCorpus.

*RoBERTa* – RoBERTa (short for Robustly optimized BERT technique) is a precise architectural clone of BERT with a larger dataset and tuned hyperparameters for pretraining that was introduced in (Liu et al. 2019). The pre-training masking approach has been changed from static to dynamic. Static masking is performed just once during pre-processing, whereas dynamic masking is performed many times during pre-processing, i.e., every sequence is masked just before feeding it to the model. Furthermore, it makes use of larger pre-training datasets. The model was trained on five English language datasets: English Wikipedia, BooksCorpus, the Stories dataset, the Open Web data, the CC-News data.

*DistilBERT* – (Tang et al. 2109) proposes distillation in neural networks, which tries to increase the speed of models. It is accomplished by using a simplified version of the BERT architecture with few parameters. It takes the original BERT's design, decreases the layers to half of the original BERT, and eliminates poolers and token embeddings to create a faster and smaller BERT for common applications. For better inference, it uses dynamic masking and overlooks the next sentence estimates. The model was trained on the same datasets as BERT, English Wikipedia, and BooksCorpus datasets.

## 4 OUTCOME OF SURVEY

This section will discuss the results for each of our research questions. It discusses the findings of primary studies that were used to generate research questions. For every study question, tables with combined outcomes as well as detailed explanations and analyses are provided. It's important to remember that different types of emotion detection tasks necessitate distinct features and methodologies. As a result, each approach is given a brief description.

### 4.1 RQ1 - What are the different Artificial Intelligence (AI) approaches used for text-based emotion detection?

There are different ways to detect emotions in text. These different ways constitute the different approaches in text-based emotion detection. Different approaches follow the different tasks, distinct processes. As a result, it is important to consider which tasks comprise which approach. The approaches utilized in various emotion detection tasks are discussed in section 3. Keyword-based, rule-based, machine learning-based, and deep learning-based techniques are commonly employed for text-based emotion detection tasks. Table 9 presents an overview of emotion models, feature extraction methods, approaches used in text-based emotion detection. Approaches that are expressly mentioned in the cited article are marked with a checkmark. The findings are organized into seven categories, each with its subcategories. The following is a list of each category and its subcategories:

*1) Emotion models* – In the text-based emotion detection literature, the discrete or categorical emotion model is widely used, which states that some emotions are more basic, more universal, or more important.

Emotion models have been extensively used in the keyword-based, rule-based, and machine learning approaches, but are less preferred in the deep learning approach.

  *2) Feature extraction methods* – Various feature extraction methods are based on lexical analysis, semantic analysis, syntactic analysis. Parts of Speech (PoS) Tagging, Bag of Words (BoW), Term Frequency Inverse Document Frequency (TF-IDF), all are the most used feature extraction methods in keyword-based, rule-based, and machine learning approaches. At the same time, word embeddings are the most preferred feature extraction method in the deep learning approach. In most deep learning-based emotion detection studies, pre-trained word embeddings Glove, Word2Vec, and FastText are employed. The table shows the most preferred word embeddings in TBED. Pre-trained word embeddings are embeddings that have been learned in one task and can be utilized to solve an identical problem. These embeddings are created by training them on big datasets, saving them, and then using them to solve other problems. Pretrained word embeddings are a type of transfer learning.

**Table 6** Comparison analysis of different pre-trained models

| Ref. | Model | Architecture | Description | Tuning | Advantages | Challenges | Trained on Datasets |
|---|---|---|---|---|---|---|---|
| Devlin et al. 2019 | BERT | Transformer, Bidirectional | Bidirectional LSTM, Uses and the Next Sentence Prediction (NSP), Masked Language Modelling (MLM) mechanisms. | Pre-Training Fine Tuning | 1) ability to handle contextual information 2) Faster Training | 1) limited to monolingual classification, 2) Fixed length of input sentences 3) Suffers from logical inference 4) computationally expensive. | BooksCorpus datasets and the English Wikipedia. |
| Howard et al. 2018 | ULMFiT | Fully Connected Neural Network, Unidirectional | Uses AWD-LSTMs | Pre-Training Fine Tuning | 1) Transfer Learning in Language Domain 2) very well on small and medium corpora 3) Prevent over-catastrophic forgetting when fine-tuning | 1) Model not trained on GLUE benchmark. 2) Unidirectional | Wikitext-103, IMDb, TREC-6, Yelp-full, Yelp-bi, DBpedia AG's news, |
| Radford et al. 2018 | GPT | Transformer, Unidirectional | to model language, a semi-supervised learning approach employed utilizing a multi-head attention layer transformer decoders. | Pre-Training Fine Tuning | 1) improved lexical robustness. 2) Outperforms various models trained on domain-specific datasets 3) no-fine tuning required for GPT-2 and GPT-3. | 1) The model's resource-intensive nature makes the pre-training step costly. 2) 2) inability to deal with dependencies that are larger than the given fixed lengths. | BooksCorpus, WebText |
| Yang et al. 2019 | XLNet | Auto-Regressive Transformer, Bidirectional. | Use two-stream self-attention and Permutation Language Model principles are used (PLM) | Pre-Training Fine Tuning | 1) The ability to extract bi-directional contextual Information 2) It removes BERT's fixed-length limitation. | 1) XLNet computationally intensive. | English Wikipedia, Common crawl, Giga 5, Books Corpus, Clueweb 2012-13 |
| Liu et al. 2019 | RoBERTa | Transformer, Bidirectional | Replication of BERT with larger corpus and tuned hyper-parameters. Uses Dynamic Masking. | Pre-Training Fine Tuning | 1) The use of more extensive pre-training data results in improved performance. 2) In downstream NLP tasks, outperforms XLNet and BERT. | 1) resource-intensive nature 2) It is computationally intensive and takes longer to complete. | English Wikipedia, Books Corpus, CC-News, Stories, OpenWebText. |
| Tang et al. 2019 | Distil BERT | Transformer, Bidirectional | Uses Early Form of BERT by reducing the of layers by a factor of 2, Uses dynamic masking. | Pre-Training Fine Tuning | 1) Language modeling capability with the pre-training 2) The new model is faster and lighter compared with BERT. | 1) Fixed length limitation | Book Corpus and English Wikipedia |

*3) TBED approaches* – TBED algorithms learn a model from text, and emotion labels, are often referred to as training data. The Model then guesses the emotion label for new, previously unknown text. Finally, the accuracy of the model is tested on a test set for which emotion labels are also available is used to determine its performance. The table 7 shows a summary of approaches used in text-based emotion detection.

1) *Deep Learning* - Deep learning models are convoluted structures that use many layers of neural networks to obtain complex information from input. In text-based emotion detection, CNN, RNN, Pre-trained deep learning models based on transformers are used. CNN uses convolutional filters to identify the pattern of data. CNN's are typically utilized in image processing applications. Still, they have recently begun to be applied in the field of natural language processing (NLP) in conjunction with other deep learning approaches. In NLP tasks, RNN is the most popular deep learning algorithm. RNNs are useful for storing sequential information. LSTM and GRU are special types of modern RNN. These modern RNN are good at learning long-term dependencies in the text. In NLP, where long-term dependencies are prevalent, Bi-LSTM and Bi-GRU, variants of LSTM and GRU, are extremely useful. Attention networks are used to concentrate on data parts that are of particular interest. Then transformer-based pre-trained models used attention with encoders and decoders to improve the extraction of relational context, eliminate the problem of long-term sequence dependencies, and allow input sequences to be delivered in parallel. Bidirectional Encoder Representations from Transformers (BERT), Generative Pre-Training (GPT), Transformer-XL, Robustly Optimized BERT pre-training Approach (RoBERTa), XLNet, and DistilBERT are popular pre-trained models. Due to its bidirectional capability, BERT has the potential to manage contextual information extraction and the fact that it trains faster and has been employed in various language modeling applications. Following the popularity of BERT, various modifications on top of BERT were developed, including RoBERTa, DistilBERT, and SpanBERT. RoBERTa is a variation of BERT that has the advantage of using a larger amount of pre-training data, which results in superior performance on a range of tasks. The concept of distillation in neural networks was suggested by DistilBERT as a technique to speed up models. DistilBERT is a language modeler that may be pre-trained to perform different linguistic modeling tasks. It significantly achieves the original BERT in terms of speed and lightness. XLNet is a language model that is self-regressive that can extract contextual information by using a permutation language model. It also removes BERT's fixed-length restriction. For predicting the next word, GPT is a good language model. The strength of GPT is improved lexical robustness. GPT-2 and GPT-3 variations do not require finetuning. The discussed deep learning algorithms are thought to be viable strategies for improving text-based emotion detection task performance.
2) *Machine learning* - The machine learning technique tackles the emotion detection

problem by implementing machine learning algorithms to classify texts into distinct emotion categories. A supervised or unsupervised ML approach is frequently used for detection. In text-based emotion detection, commonly supervised machine learning algorithms are implemented. Classical machine learning techniques are still frequently utilized in all types of emotion detection tasks similar to text classification. Even though deep learning is a rising field, for several tasks, classical machine learning algorithms perform as well as or better than deep learning approaches, especially on smaller datasets. Algorithms for supervised machine learning that have been around for a long time are Naive Bayes (NB), Support Vector Machines (SVM), Random Forest (RF), Neural Networks (NN), k-Nearest Neighbor (k-NN), Logistic Regression (LR), and Decision Trees (DT). SVM, NB, and RF are more popular machine learning algorithms as compared to others.
3) *Rule-Based Approach* – In rule-based learning, grammatical and logical concepts are used to construct the rules for detecting the emotions from the text. Rule-based learning is built on exploiting information to understand data in a meaningful way.
4) *Keyword Based Approach* - The traditional approach to TBED is the keyword-based approach. It looks for words in documents that indicate different emotions in humans. This method does not require any learning data because words are specified in a dictionary/dataset. The goal is to locate occurrences of search words at the sentence level in a given document. Once the keyword has been located inside the sentence, the sentence is labeled according to the dataset.

## 4.2 RQ2 - What application domains have been adopted in text-based emotion detection?

Public thoughts, opinions, views provide us with valuable insights. Detecting the emotions from these opinions on the online digital platform will help to learn users, and it has a wide range of applications domains. This task of analyzing emotions can help in a wider range of application domains, including management and marketing, healthcare, education, finance, public monitoring, dialog systems/chatbots, social media mining.

We used the methodological choices on these application domains in the realm of text-based emotion recognition to survey articles on online social media, review systems, and conversational agents in section 3. Table 8,10,11 shows the different methods and techniques used in different text-based emotion detection applications.

Online social media (OSM) applications have become the universal applications on the Internet. These are rapid data generation applications making it subtle to study these data. Text-based emotion detection has been widely used in online social media for a variety of tasks. These tasks include detecting sarcasm, irony, tracing the flow of emotions in social media, determining personality traits, depression detection, etc. Table 10 shows applications in social media on which research has been done in TBED includes Twitter, Facebook, YouTube. Other topics like emotion detection from news articles, behavioral analysis, crime detection, business intelligence, recommendation systems, stock market prediction, politics have also been studied. The domain of online social media has preferred machine learning classifiers: SVM, k-NN, random forest, decision tree, Naïve

Bayes, and logistic regression, whereas Bi-LSTM, Bi-GRU techniques from deep learning are the most used.

Customer reviews on products or services can provide sentiment or emotions towards various features or aspects. Therefore, the analysis of these reviews can extract meaningful insights, which will help businesses maintain customer relationships, gain feedback from users to help in early product designs, and thus improve the businesses.
In the domain of review systems, some work has been done in product review, movie review, hotel recommendations systems, etc. Techniques used in review systems are naive Bayes, SVM, maximum entropy, k-means clustering, convolutional deep belief networks (CDBN), expectation-maximization semi-supervised algorithm, and syntactic rules with polarity determination.

Conversational agents, often known as chatbots, are beneficial in a variety of industries, including education, commerce and eCommerce, health, and entertainment. The most crucial goal for chatbot users is productivity, while other factors such as enjoyment, social factors, and novelty engagement are also important. Table 11 shows different applications areas where conversational agents can be used.

## 4.3 RQ3 - What are the different datasets available for research purposes in text-based emotion detection, and which domains have been addressed in the available data sets?

Labeled or annotated tailored datasets are available for text-based emotion detection research. Existing datasets were favored by researchers, or they built their own datasets based on the needs of studies in certain application domains. Here the authors have identified some understanding of datasets with different perspectives.

The authors have inferred from table 12, online social media, customer/product review systems, conversational/dialogue systems, and news articles are the application areas addressed by currently available datasets in text-based emotion detection. Most of the datasets have been created to study the emotions/sentiments on online social media. Many datasets are imbalanced in terms of emotion/sentiment labels. All online available datasets are annotated/labeled. The majority of the datasets are openly accessible and can be downloaded. The presence of a checkmark in the table indicates that specifically mentions datasets are balanced or not, annotated or not, access free or not, and multi-modal or not. In contrast, most datasets are not multi-modal. The important inference is that the majority of datasets are multi-labeled datasets that can be used to detect text emotion.

Emotion detection from text is considered a multi-class classification problem. In text-based emotion detection, researchers use machine learning or deep learning methods, which require huge amounts of data. On the other hand, supervised learning algorithms provide good results with annotated datasets. But the main problem with annotated datasets is, very few annotated datasets are available. Although in Table 12, all datasets studied are annotated, but these datasets are very less in number.

**Table 7** Summary of text-based emotion detection approaches

| Ref | Approach | Description | Merits | Demerits |
|---|---|---|---|---|
| Tao J. 2004, Ma C. et al. 2005 Perikos et al. 2013 Shivhare et al. 2015 | Keyword Based | Based on finding the occurrence of a word and comparing it with annotated labels in the dataset. | - Simple method<br>- Most used approach. | - The emotion keyword does not always match.<br>- Emotion can present without emotional words. |
| Lee et al. 2010, Liu et al. 2017 Udochukwu et al. 2015 | Rule-Based | Based on grammatical and logical rules from text to interpret the information. | - Simple approach.<br>- Rules are easily created for small sets. | - Rules creation for large data sets complex task.<br>- Lack of linguistic information affect the performance |
| Aman et al. 2007, Ghazi et al. 2010 Bruyne et al. 2018, Suhasini et al. 2020 Singh et al. 2019, Allouch et al. 2018 Jonathan et al. 2017 | Machine Learning-Based | Based on the ability to learn from experience and develop. | - Widely implemented<br>- Better detection results. | - Not robust.<br>- Not explicitly extract semantic information. |
| Baziotis et al 2017, Ezen and Can 2018 Basile et al. 2019, Xiao 2019 Shrivastava et al. 2019, Polignano et al. 2019 Rathnayaka et al. 2019 Bian et al. 2019, Jose et al. 2020 Mohammad et al. 2021, Xia and Zhang 2018, Diogo 2021, Taiao et al. 2020 | Deep Learning Based | Based on to learn without human intervention from data. | - More robust<br>- Extract deeply hidden details. | Need large data to train the system. |

**Table 8** Summary of Review systems in different application areas

| Ref. | Application Area in Review Systems | Description | Techniques/ Methods used | Dataset/data utilized | Challenges |
|---|---|---|---|---|---|
| Bai et al. 2019 | Product Design | Utilizing user experience/feedback from online customer reviews for product design. | PoS tagging, Bootstrapping technique, feature ranking approach, Expectation-maximization semi-supervised algorithm | Amazon | The proposed model does not directly link to emotions or user sentiment at the detailed level. Limited domain knowledge and other online product review websites are not considered. |
| Monireh et al. 2018 | Product Review | Proposed an automatic tool to analyses customer reviews of a product and determine customer sentiments about that product. | Syntactic rules with polarity determination, Lexicons used – SentiStrength, Bing | Pros/Cons Product review for digital camera | Cannot analyze the service/movie reviews. Language dependent. |
| Xiang et al. 2017 | Hospitality and Tourism online review system | Examines the information quality of three major online review sites in relation to online reviews about the entire hotel population in New York City. | Latent Dirichlet Allocation, Elbow Method, Naïve Bayes, TF-IDF | TripAdvisor, Expedia, and Yelp | Data used are destination-specific. A very specific case discussed |
| Cambria et al. 2020 | Product Review | Proposed domain adaptation using genetic programming for Amazon product reviews for a variety of products and languages. | Domain Adaptation using genetic programming, Semi-supervised convolutional deep belief network (CDBN) | Multi-domain Amazon dataset | Variance is high during the search. |
| Topal et al. 2016 | Movie Review | Presented the notion of emotion maps centered on machine learning techniques | k-means clustering, emotion heat maps. | IMDb movie dataset | Very few movies were considered for the study. |
| Teja et al. 2021 | Customer care center Product Review | Predict customer satisfaction following interactions with a corporate call center. | GRU, CNN, GNN | Amazon | Human-in-the-loop / active learning approaches can improve system |

**Table 9** Overview of emotion models, feature extraction methods, approaches used in TBED with is subcategories

| Ref. | Emotion models Used | | | Feature Extraction Methods | | | | | Word Embeddings | | | | | Text Pre-processing | Keyword-based | Rule-Based | Machine Learning-Based | | | | | | | Deep Learning-Based | | | | | | | | |
|---|---|---|---|---|---|---|---|---|---|---|---|---|---|---|---|---|---|---|---|---|---|---|---|---|---|---|---|---|---|---|---|---|
| | Discrete | Dimensional | Componential | BoW | TF-IDF | PoS | NER | Negation | Glove | Word2Vec | FastText | GoogleEmb | Not Mentioned | | | | SVM | NB | RF | k-NN | DT | LR | ANN | CNN | Bi-LSTM | GRU | BERT | RoBERTa | DistilBERT | XLNet | GPT | ULMFiT |
| Tao 2004 | ✓ | | | | | ✓ | | | | | | | | ✓ | ✓ | | | | | | | | | | | | | | | | | |
| Ma et al. 2005 | ✓ | | | | | ✓ | | ✓ | | | | | | ✓ | ✓ | | | | | | | | | | | | | | | | | |
| Perikos et al. 2013 | ✓ | | | | | ✓ | | ✓ | | | | | | ✓ | ✓ | | | | | | | | | | | | | | | | | |
| Shivhare et al. 2015 | ✓ | | | | ✓ | | | | | | | | | ✓ | ✓ | | | | | | | | | | | | | | | | | |
| Lee at al. 2010 | ✓ | | | | | ✓ | | | | | | | | ✓ | | ✓ | | | | | | | | | | | | | | | | |
| Udochukwu et al. 2015 | ✓ | | | | | ✓ | | | | | | | | ✓ | | ✓ | | | | | | | | | | | | | | | | |
| Liu et al. 2017 | | | | ✓ | ✓ | ✓ | | | | | | | | ✓ | | ✓ | | | ✓ | | ✓ | | | | | | | | | | | |
| Aman et al. 2007 | ✓ | | | | | | | | | | | | | ✓ | | | ✓ | ✓ | | | | | | | | | | | | | | |
| Ghazi et al. 2019 | ✓ | | | ✓ | | | | | | | | | | ✓ | | | ✓ | | | | | | | | | | | | | | | |
| Bruyne et al. 2018 | ✓ | | | ✓ | | ✓ | | | | ✓ | | | | ✓ | | | ✓ | | ✓ | | | ✓ | | | | | | | | | | |
| Suhasini et al. 2020 | | ✓ | | | | | | ✓ | | | | | | ✓ | | ✓ | | ✓ | | ✓ | | | | | | | | | | | | |
| Singh et al. 2019 | ✓ | | | | | ✓ | | | | | | | | ✓ | | | ✓ | | | | | | | | | | | | | | | |
| Allouch et al. 2018 | | | | | ✓ | | | | | | | | | ✓ | | | ✓ | ✓ | | ✓ | | ✓ | | | | | | | | | | |
| Jonathan et al. 2017 | | | | | | | | | | | | ✓ | | ✓ | | | ✓ | | | | | | | | | | | | | | | |
| Baziotis et al. 2017 | | | ✓ | | | | | | ✓ | | | | | ✓ | | | | | | | | | | | ✓ | | | | | | | |
| Ezen et al. 2018 | | | | | | | | | ✓ | | | | | ✓ | | | ✓ | ✓ | ✓ | | ✓ | | | | | ✓ | | | | | | |
| Basile et al. 2019 | ✓ | | | | | | | | | | | | ✓ | ✓ | | | | ✓ | | | | | | ✓ | ✓ | | ✓ | | | | | |
| Shrivastava et al. 2019 | | | | | | | | | | ✓ | | | | ✓ | | | | | | | | | | | ✓ | | | | | | | |
| Xiao 2019 | | | | | | | | | | | | | ✓ | ✓ | | | | | | | | | | | ✓ | | ✓ | | | ✓ | ✓ | |
| Rathnayaka et al.2019 | | | | | ✓ | | | | ✓ | | | | | ✓ | | | ✓ | | | ✓ | ✓ | | | | | ✓ | | | | | | |
| Bian et al. 2019 | | | | | | | | | | | | | ✓ | ✓ | | | | | | | | | | ✓ | | | | | | | | |
| Mohammad et al. 2021 | | | | ✓ | | | | | ✓ | | | | | ✓ | | | | | | | | | | | ✓ | ✓ | | | | | | |
| Jose et al. 2020 | | | | | | | | | | ✓ | | | | ✓ | | | ✓ | | | | | | | | ✓ | | ✓ | | | | | |
| Xia et al. 2018 | | | | | | | | | ✓ | | | | | ✓ | | | ✓ | | | | | | | | ✓ | ✓ | | | | | | |
| Diogo 2021 | | | | | | | | | ✓ | ✓ | ✓ | ✓ | | ✓ | | | ✓ | | | | | | | | ✓ | ✓ | ✓ | ✓ | ✓ | ✓ | | |
| Taiao et al. 2020 | | | | | | | | | | | | | ✓ | ✓ | | | ✓ | ✓ | ✓ | | | | | | ✓ | ✓ | ✓ | | | | | |
| Polignano et al. 2019 | ✓ | | | ✓ | ✓ | | | | ✓ | | ✓ | ✓ | | ✓ | | | ✓ | | | | | | | | ✓ | | | | | | | |

**Table** 10 Summary of online social media in different application aeras

| Ref. | Application Area in Online social media | Description | Techniques/ Methods used | Dataset/data utilized | Challenges |
|---|---|---|---|---|---|
| Kumari et al. 2021 | Behavioral Analysis | A proposed text-based cyberbullying detection system in social media | Stacked denoising auto-encoder, Word embeddings | Twitter and myspace | Not considered word order in messages which can further improve the robustness of the learned representation. |
| Wang et al. 2017 | Business intelligence | Developed a methodology to identify a group of people who are strongly linked and emotionally engaged in their feelings about a product or service. | Linear threshold and independent cascade diffusion models. | Flixster.com movie review dataset | Accuracy needs to be improved of community sentiment detection methods. Need to consider the demographic characteristics of the most valuable customers. Not observed dynamic customer sentiments from social media. |
| Song et al. 2017 | Crime detection | Presented Social Spam detection system | Labeled latent Dirichlet allocation (L-LDA) | You-tube | Aside from topic-based elements, other high-level features must be learned. Need to explore other application domains besides YouTube. Need to improve the quality of topic-based features. |
| Gupta 2021 | Epidemics | The weather, which has been discussed, impacts the virus's transmission over the world. | Machine learning algorithms SVM, NB, RF, and K-means clustering with Topic modeling. | Twitter | Mislabeled data. Not considered the magnitude of tweets while annotating the data. Need to improve the performance of the classifier for sarcasm. |
| Sun et al. 2015 | Recommendation systems | Recognizing the appropriate class of friends to offer recommendations. | Bi-clustering Algorithm, SVM | Del.icio.us | Need to improve the performance. Time-series information, place information, and the dynamic variations among users are not considered for recommendations. |
| Douiji et al. 2016 | Sentiment, opinion, and emotion analysis | Created a framework to handle the complexities of chat writing styles and determine user sentiment. | Latent Semantic Analysis (LSA), Probabilistic Latent Semantic Analysis (PLSA), Non-negative Matrix Factorization (NMF) | You-tube | NA |
| Pool et al. 2016 | | Proposed system to classify emotions in Facebook posts | SVM | Facebook | There is a lot of variation in ratings for the same emotion among datasets. |
| Sailunaz et al. 2019 | | Calculate user's influence scores using user and tweet information to classify the emotion and sentiment. | Naïve Bayes | Twitter | Not considered abbreviations, emoticons, short-hand text. The system is static, with a narrow range of data acquired in less span of time and a major amount of missing information on individuals. |

**Table 11** Summary of Conversational Agent systems in different application areas

| Ref. | Application Area in Conversational Agents | Description | Techniques/ Methods used | Dataset/data utilized | Challenges |
|---|---|---|---|---|---|
| Basile et al. 2019 | English user-chatbot interaction in the Indian chat room. | To recognize emotional responses between a human and a Chatbot. | Bi-LSTM with deep attention mechanism, USE, BERT, Ensemble Model (RF, SVM, LR) | SemEval-2019 | very unbalanced distribution of data. Accuracy is not up to the state of art systems |
| Adikari et al. 2019 | Industrial chatbots | Proposed cognitive model to detect emotions in industrial chatbots. | Markov chains, word embedding, and NLP using emotion vocabulary | service dialog dataset | System performance can be improved. |
| Thomas 2016 | E-business chatbots | Proposed a chatbot that responds to people using a frequently asked questions (FAQs) database. | Artificial Intelligence Markup Language (AIML) and Latent Semantic Analysis (LSA). | The frequently answered questions from the internet for the e-business domain | Only designed for FAQ's. General questions are not considered. |
| Xue et al. 2018 | Call center customer support agent. | Developed a neural-based conversational solution to boosts the effectiveness of customer support agents. | Bi-LSTM with an attention mechanism | Raw Slack channel data | Less labeled data used for training the intent classification model |
| Fadhil et al. 2019 | Healthcare | CoachAI is a conversational agent-assisted health coaching system that supports the delivery of health interventions to individuals or groups. | Support Vector Machine (SVM), Finite-state Machine architecture. | An ambulatory health care clinic data was utilized. | Activities suggested are not clinically validated and guided by a health care provider. |
| Hien et al. 2018 | Education | Discussed smart learning contexts and demonstrated a chatbot that, on behalf of the academic staff, responds to a question from students regarding the services provided by the education system. | Named Entity Recognition (NER), Text Classification | Student Data was utilized from the Ho Chi Minh City University of Science. | System created for a specific application domain. There is no knowledge structure between the user inputs and the responses generated. Only dealt with the user's context to a limited extent. |

The second problem is that creating these annotated datasets is based on a particular emotion model, so the dataset displays only those emotions present in that specific emotion model. Again, the third problem is creating an annotated dataset with manual labeling is a time-consuming and error-prone task. Datasets for emotion detection are nothing but user-generated text, and it completely depends on the domain from which data is generated and the source it has generated. TBED has spread the wing into application domains like online social media and product/customer reviews. If we study, the writing style of online social media is very informal, with lots of emojis, hashtags, etc. In contrast, product and customer reviews are formal and explicate a particular product/service. Different online datasets available to researchers have been discussed in section 3.3. These datasets have some application domains adopted in them. When it comes to evaluating any learning model on datasets, it performs well on specific application domains, but performance gets degraded when tested on different application domains. The domains of the adopted datasets with other details are depicted in Table 12.

## 4.4 RQ4 - What are the challenges and open problems with respect to text-based emotion detection?

The authors have presented different datasets and broad categories of text-based emotion detection applications from the literature survey. Though our study highlights the different revelations of text-based emotion detection, there are still some challenges and research directions that need to be addressed to align emotion detection with the requirements of the real world. Table 13 shows the challenges identified. The following are some of the research challenges and drawbacks in this field:

A major challenge in text-based emotion detection is related to datasets. Only a few datasets are existing for research, and the majority of them are imbalanced datasets. These datasets were created for specific analysis; therefore, application areas are dependent. So, domain dependency of datasets is one of the challenges in text-based emotion detection. Language dependency is also one of the major challenges. Most of the research was done in English text corpora. Other languages such as Chinese and Arabic are preferred after English. So, building multilinguistic systems is a challenging task.

The second challenge is the performance of machine learning and deep learning techniques. These techniques have been used extensively in text-based emotion recognition. Though, many machine learning algorithms need labeled datasets, that is a tedious task and is reliant on the efficacy of humans, lowering the performance of machine learning techniques. Deep learning approaches, on the other hand, are emerging, but they are complex procedures that can keep word order and syntactic patterns. Deep learning approaches still pose certain issues, like a vast amount of data required for training. Text-based emotion detection is a multi-class problem where the efficiency and performance of deep learning and machine learning techniques can be degraded. The third challenge in TBED is the quality of text used in online media. Incomplete information, typing mistakes, slang words, short texts, emojis, sarcasm, irony, harmony, etc., all these inconsistencies are used in languages on online media by users. These inconsistencies make emotion detection a challenging task. Other further challenges related to text semantics are the inability to recognize implicit emotions, unable to extract the semantic data, classifying emotions according to their intensities, mislabeling emotions. These, too, are important challenges that need to be focused on by researchers.

**Table 13** Overview of Datasets

| Dataset | Application Domain | Granularity | Size | Balanced/ Unbalanced | Annotated/ Not annotated | Access Type Free/not free | Multi Modality | Emotion Labeled |
|---|---|---|---|---|---|---|---|---|
| **ISEAR** Scherer et al. 1994 | Social Media | Blogs | 7666 Sentences | ✓ | ✓ | ✓ | ✗ | Anger, disgust, fear, sadness, shame, joy, and guilt |
| **Alms' Affect data** Alm et al. 2008 | -- | Story Tales | 15302 sentences | ✗ | ✓ | ✓ | ✓ | Disgust, anger, fear, sadness, joy, positive surprise, and negative surprise |
| **SemEval 2016** Rosenthal et al. 2019 | Social Media, | Blogs, Tweets | 22761 | ✗ | ✓ | ✓ | ✗ | Positive, Negative, Neutral |
| **SemEval 2007** Rosenthal et al. 2019 | Discussion Forums | News articles | 1,250 | ✗ | ✓ | ✓ | ✗ | Anger, disgust, fear, sadness, joy, and surprise |
| **SemEval 2019** Rosenthal et al. 2019 | Conversations | Dialogues | 38424 | ✗ | ✓ | ✓ | ✗ | Happy, sad, angry, other |
| **Crowdflower** https://data.world/crowdflower/ sentiment--in-analysistext | Social media | Tweets | 40001 tweets | ✗ | ✓ | ✓ | ✗ | Boredom, anger, empty, fun, enthusiasm, happiness, love, hate, neutral, sadness, relief, worry, surprise. |
| **Emobank** Buechel et al. 2017 | Social media | Blogs | 10062 | ✓ | ✓ | ✓ | ✗ | Valence, Arousal, Dominance |
| **Aman** Aman et al. 2007 | Social media | Blogs | 4266 Sentences | ✗ | ✓ | ✗ | ✗ | Sadness, happiness, anger, disgust, surprise, fear, no emotion, and mixed emotion |
| **The Valence and Arousal Facebook Posts** Preotiuc et al. 2016 | Social media | Facebook | 3120 posts | ✓ | ✓ | ✓ | ✗ | Valence, Arousal |
| **MELD** Poria et. al. 2018 | Conversations | Dialogues/ utterances | 15141 | ✗ | ✓ | ✓ | ✓ | anger, disgust, fear, joy, neutral, sadness, and surprise, neutral |
| **CBET** Ameneh et al. 2015 | Social media | Tweets (Twitter) | 76,860 tweets | ✓ | ✓ | ✓ | ✓ | Joy, anger, love, surprise, fear, sadness, thankfulness, guilt, and disgust |
| **GoEmotions** Dorottya et al. 2020 | Social media | Blogs (Reddit) | 58009 | ✗ | ✓ | ✓ | ✓ | admiration, amusement, approval, anger, annoyance, curiosity, caring, confusion, desire, disgust, disappointment, disapproval, excitement, embarrassment, fear, grief, gratitude, joy, love, sadness, nervousness, optimism, pride, remorse, realization, relief, surprise. |
| **Amazon** Blitzer et al. 2007 | Review systems | Product review | 8000 reviews | ✓ | ✓ | ✓ | ✓ | Positive, Negative |
| **IMDB** https://datasets.imdbws.com/ | Review Systems | Movie Review | 50000 reviews | ✓ | ✓ | ✓ | ✗ | Positive, Negative |
| **The Movie Review** Pang et al. 2002 | Review Systems | Movie Review | 10,662 sentences | ✓ | ✓ | ✓ | ✗ | positive, negative, or neutral |

**Table 12** Challenges identified through literature survey

| Problem area | Challenges in text-based emotion detection | Future Directions |
|---|---|---|
| Datasets | Lacking labeled/annotated datasets. | Domain Adaptation<br>Transfer Learning<br>GAN<br>Multi-modality in text and data sources |
| | Imbalanced datasets | |
| | Domain dependent datasets | |
| | Language dependent datasets | |
| | Mislabeled emotions | |
| Accuracy | Several techniques are not robust | Auto-encoders<br>Ensemble methods |
| | Accurateness of current systems | |
| Quality of text | Incomplete information, typing mistakes | Multitask Learning System<br>Transfer learning<br>Pattern-based approach. |
| | Slang words, short texts, emojis | |
| | Detection of sarcasm, irony, harmony | |
| Semantics | Inability to recognize implicit emotions | Pre-trained word embeddings<br>Best-Worst Scaling (BWS) Annotation Scheme.<br>Deep Learning algorithms |
| | Unable to extract the semantic data | |
| | Categorizing emotions considering their intensities | |

## 5 Future Directions in Text-Based Emotion Detection

Techniques or methodologies in TBED have witnessed huge evolutions over the past several years. Combining keyword-based approaches with n-hidden layer-based deep learning techniques, for example. Artificial Intelligence advances in recent years have bolstered text-based emotion detection algorithms. Some of the outstanding difficulties in text-based emotion detection have been addressed by AI-based technologies such as data augmentation (generative adversarial networks), transfer learning, explainable AI, autoencoders, adversarial machine learning, and different learning styles. Figure and table showcase several challenges and their solutions using these methods with references.

### 5.1 Data Augmentation

In Natural language processing (NLP), many tasks like text classification to question answering require vast training data to improve the performance of the model. However, acquiring and annotating additional data can be an expensive and time-consuming process in order to generate a large amount of training data. In this scenario, data augmentation techniques can be applied. Data augmentation is a machine learning technique that artificially generates additional synthetic training data through label preserving transformations (Bayer et al. 2021). Though data augmentation techniques have achieved great success in computer vision applications, they are powerful in NLP applications. Data augmentation is achieved by thesaurus, word embeddings, back translation, replacing synonyms, text generation, etc. In (Bayer et al. 2021), the authors proposed a novel text generation method to increase the performance of short and long text classifiers using data augmentation techniques. The authors used GPT-2 355 million parameter model for text generation in the implementation. The authors in (Shi et al. 2020) developed an efficient data augmentation algorithm for labeled sentences called AUG-BERT for text classification. A deep neural network-based text classifier based on data augmentation and semi-supervised learning built using free-text obtained from a survey on sleep-related issues (Shim et al. 2020). One of the techniques trending for data augmentation is GAN, explained in detail as follows –

*Generative Adversarial Networks (GAN)* – The lack of balanced data over all labels in the data is one of the most critical issues in text-based emotion detection. Class imbalance causes over-representation of majority classes, poor model performance, and classification models may disregard or treat small classes as noise. GAN is the most recent technique for dealing with imbalance. Figure 15 shows the architecture of the generative adversarial network. A generative adversarial network (GAN) is a new framework that sets two neural networks against one other (thus the term "adversarial") to create new, synthetic data instances that look like actual data. The discriminator evaluates whether each instance of data it examines corresponds to the actual training dataset or not, while the generator generates new data instances. (Gray et al. 2019) proposed spamGAN, a generative adversarial network for detecting opinion spam that uses both labeled and unlabeled data. SpamGAN outperforms state-of-the-art GAN-based text classification algorithms, according to the authors. TripAdvisor data was used in the experiments to show that spamGAN outperforms existing techniques when labeled data is limited. GANs have recently been suggested by a number of experts for unlabeled or imbalanced data (Li et al. 2020, Croce et al. 2020).

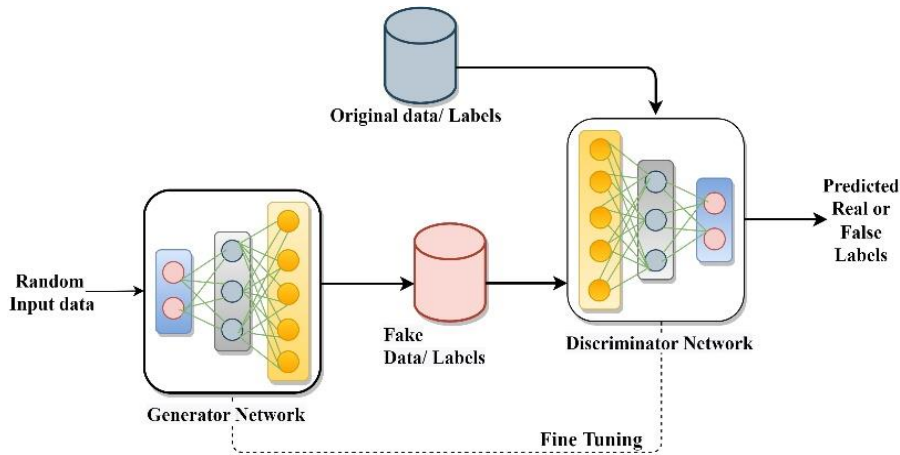

**Fig. 15** Architecture of Generative Adversarial Networks

## 5.2 Transfer Learning

Data collection and labeling are time-consuming and frequently demand expert knowledge. Transfer learning is an excellent technique to handle difficulties like data scarcity and a lack of human labeling. Transfer learning allows you to reduce the amount of labeled data you need and take advantage of a previously trained model to increase performance. The knowledge of the source domain can be transferred to separate but related target domains using transfer learning. A set of data containing adequate data samples, a large number of labels, and relatively high quality is referred to as a source domain. A target domain's data, on the other hand, may contain a small number of samples, limited amounts of data, or no labels at all, as well as being noisy. Transfer learning strategies try to increase the learning of the target task using information from the source domain when given a source and a target (Feng and Chaspuri 2019). Figure 16 gives a closer look at transfer learning. Transfer learning for emotion recognition via cross-lingual embeddings was proposed by the authors in (Zishan et al. 2020). Their developed a deep transfer learning framework using CNN and Bi-LSTM efficiently transfers the captured relevant information through the shared space of English and Hindi. (Omara et al. 2019, Devamanya et al. 2020, Yasaswini et al. 2021) proposed using a transfer learning approach for the scarcity of data.

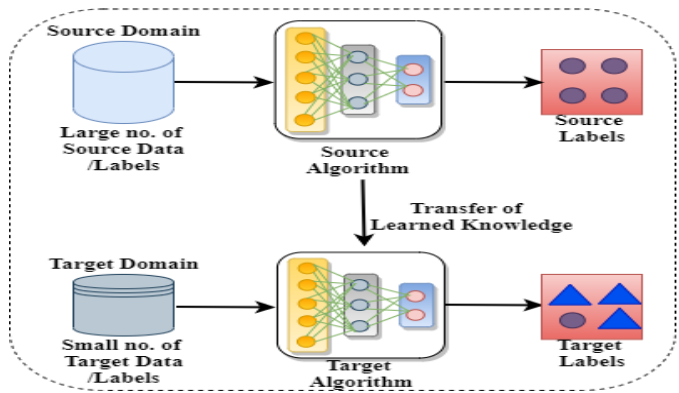

**Fig. 16** A closer look at transfer learning

## 5.3 Explainable Artificial Intelligence (XAI)

Deep learning and machine learning models have drawback of not being human interpretable. These models suffer the major problem of lack of transparency. The interpretability of AI systems is the topic that is currently generating interest in the scientific community. The non-linear and nested structure of deep neural networks, which enhances prediction model accuracy, makes it difficult to understand how information flows from a human-understandable input such as an image or a piece of text to a predicted output. Deep neural networks are regarded as "black boxes" in this sense. Explainable AI proposes that models built-in machine learning or deep learning should explain their decision-making process and actions. Very little work has been done in this area. Figure 17 presents the architecture of XAI. The

Authors in (Zucco et al. 2018)] highlighted the need for explainable sentiment analysis for medical applications in order to extract human-reliable knowledge related to opinions and emotions from the user-generated text information. In (Turcan et al. 2021) authors discussed detecting psychological stress in online posts using a more explainable black-box model. (Nawshad et al. 2021) discussed Depression Symptoms Detection (DSD) from text using Explainable Zero-shot modeling.

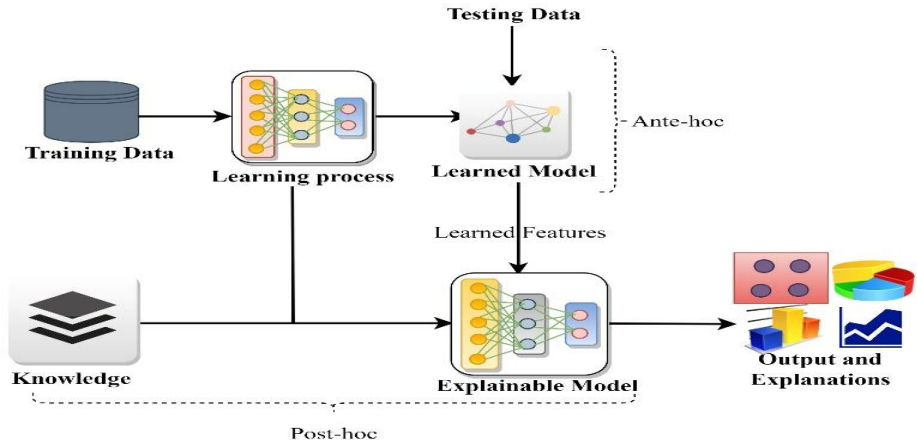

**Fig. 17** Explainable Artificial Intelligence (XAI)

## 5.4 Auto-Encoders

These are a kind of neural network where feature vectors are transferred into an arbitrary higher or lower dimensional space during the encoding phase, allowing the original feature vector to be rebuilt with minimal reconstruction error in a subsequent decoding phase. Figure 18 shows the representation of auto-encoders. Autoencoders can deal with missing data, improve system performance in noisy environments, and be trained unsupervised. Denoising autoencoders, a form of autoencoder, have been employed in domain adaptation, which involves training a model on one labeled corpus and then applying it to a separate unlabeled corpus with a different context. In (Hesam et al. 2017) the authors used stacked denoising autoencoders (SDAs) to perform sentiment recognition on a multitude of domains and languages (Kumari et al. 2021) have used autoencoders to construct a deep learning model for detecting distinct stages of violence (indirect, direct, and no aggression) from posts on social media in a multi-lingual setting.

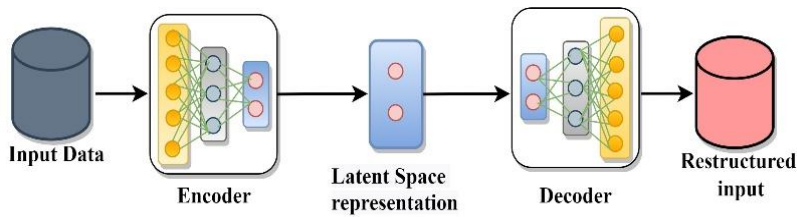

**Fig. 18** Autoencoders

### 5.5 Adversarial Machine Learning (AML)

In numerous Natural Language Processing (NLP) tasks, many machine learning and deep learning models have had remarkable success. On the other hand, machine learning and deep learning models, are increasingly vulnerable to adversarial examples, according to a growing number of research publications. Adversarial machine learning (AML) is an emerging research area that brings together the best machine learning practices, robust statistics, and computer security (Deldjoo et al. 2021). The act of deploying attacks towards machine learning-based systems and making systems robust to attacks is known as Adversarial Machine Learning (AML). In adversarial machine learning, inputs are modified by creating small perturbations to cause a malfunction in the target model, which outputs incorrect results. Figure 19 shows the adversarial machine learning model. The aim is to exploit the weaknesses or vulnerabilities of the machine learning model. As a result, the model's usefulness can significantly be reduced. The authors in (Wei et al. 2019, Basemah et al. 2020) surveyed adversarial attacks on deep learning models in natural language processing. While in (Deldjoo et al. 2021), discussed attack/defense strategies for adversarial recommender systems, generative adversarial networks in detail. (Alsmadi 2021) examines features of adversarial machine learning, specifically in text analysis and generation, and outlines key research advances in the field, including GAN algorithms, models, attack types, and countermeasures.

### 5.6 Learning Styles

Machine learning is the learning from experience, that is, acquiring skills or knowledge from experience. It means creating useful information from past data. There are many different learning styles depending on how algorithms use multiple layers to extract increasingly higher-level features from the raw input. Figure 20 shows different learning styles. These learning styles are becoming the advanced trends in text classification or NLP applications. Table 14 gives the summary of learning styles.

*1) Co-learning* – Co-learning is based on a co-training concept. The co-training approach was first introduced by (Blum and Mitchell 1998). Their approach proposed that multiple classifiers can learn from each other using different features of data. First, all classifiers are trained on a small dataset (labeled) in a completely supervised method in co-learning. Then each trained classifier is tested on unlabeled data. This automatically

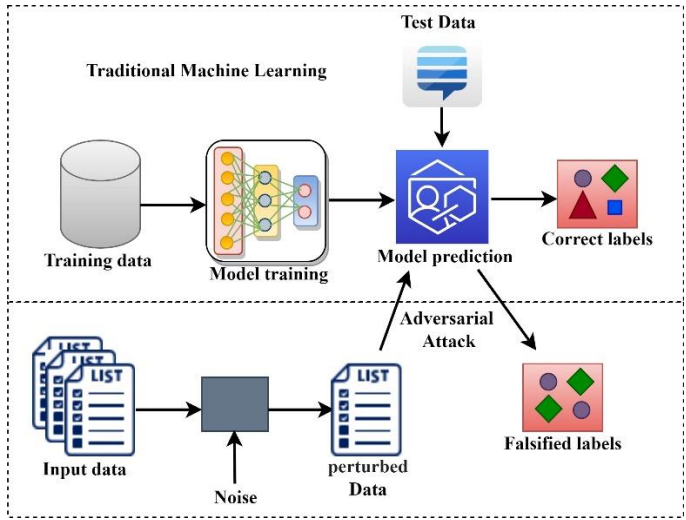

**Fig. 19** working of Adversarial Machine Learning (AML)

labeled data, by each classifier, is then used to re-train the other classifiers in a semi-supervised method. Co-learning is used when lack of training data or low/unavailability of labeled data. This learning strategy can improve the generalization capacity of classifiers with learned features, resulting in better classification models. The authors developed a unique technique for multi-domain text categorization called dual adversarial co-learning (Wu et al. 2020). Their method uses shared-private networks to extract both domain-invariant and domain-specific features, then uses the extracted features to train two classifiers. Finally, these classifiers are co-learned on unlabeled data in an adversarial manner. With the learned characteristics, the authors intended to improve the generalization capacity of the classifiers. The credibility of Arabic web blogs is assessed in (Helwe et al. 2019) by utilizing a deep co-learning approach. The authors showed a lack of training data in the Arabic web blogs co-learning approach performed well with other baseline methods.

*2) Multi-task learning* - Multi-task learning is a method in which the model is trained simultaneously on many tasks rather than separately on each task. The goal is to create a generalized model that can tackle a variety of tasks. Specifically, information from the training signals of related tasks is utilized in multi-task learning. To learn several linked tasks, use share representation. By sharing representations between related tasks, Multi-Task Learning (MTL) facilitates model generalization. There are two mechanisms in multi-task learning: 1) (a) the soft-sharing mechanism that applies a task-specific layer to different tasks. (b) the hard-sharing mechanism that utilizes a shared feature space to extract features for different tasks. Multi-task learning has the potential to make a model more robust in the limited dataset. The authors of (Zhang et al. 2021) created a multi-task learning framework for sentiment analysis in several application domains based on a hard-sharing approach. (Akhtar et al. 2020) proposes a multi-task learning framework for finding and classifying aspect elements in a unified model. The authors used a Bi-directional Long Short-Term Memory (Bi-LSTM) and a Convolutional Neural Network (CNN) network in a cascade for collaborative learning to extract aspect terms and classify aspect sentiment in a multi-task framework.

*3) Zero-shot learning/Few shot learning* – In most machine learning classification models, all data labels are evaluated in the course of training, but, in some cases, it is necessary to categorize instances whose classes are not observed at all or only a few times during training. Issues related to target classes are such as a small number or a large number of classes, expensive class labeling, or changing classes over time. At the time of the testing when there is no labeled example, it is known as zero-shot learning; at the time of the testing when there is only one labeled example, it is termed one-shot learning; and at the time of the testing when there are a few labeled instances, it is referred as few-shot learning. In (Thakkar et al. 2021), the authors use zero-shot and few-shot learning to do a cross-lingual sentiment analysis of news articles. The authors evaluated their system on Croatian and Slovene datasets using single-task and multi-task environments. A multi-label few-shot learning system was proposed in (Hu et al. 2020) for Aspect category detection (ACD) in sentiment analysis.

*4) Meta-learning* – To solve a novel problem, the meta-learning framework collects a large number of tasks, each of which is treated as a training example, and then trains a model to adapt to all of those training tasks. Finally, this model should perform well in the new assignment. Meta-Learning is an algorithm that trains itself by learning to perform many tasks on training data before being tested on new ones. Because it enhances data efficacy, allows learned information transfer, and facilitates unsupervised learning. Meta-learning is crucial for both task-specific scenarios and task-agnostic. It is a method of addressing challenges that makes advantage of the experience gained while completing related tasks. The major objectives are to expedite the learning process and improve quantitative model performance. (Madrid et al. 2019) highlighted the problem of impulsive recognizing the category of text classification task using meta-features. The authors provided a collection of 73 meta-features that were tested in 81 data sets related to five different types of tasks. (Yin 2020) surveyed meta-learning for a few shot NLP applications.

*5) Reinforcement learning* – It is a machine learning technique where an agent is trained in an interactive environment through reward and punishment mechanisms using feedback from its own actions and experiences. The agent is rewarded for correct actions and punished for the wrong ones. By learning from own actions and experiences, the agent tries to minimize wrong actions and maximize the right ones. Reinforcement learning is used in which a sequence of decisions is needed. Most reinforcement learning environments use Markov decision processes as a mathematical framework. In natural language processing (NLP) applications, reinforcement learning is used for text summarization, Q&A, dialogue systems, machine translation, etc. In (Duo et al. 2020), the authors proposed a framework to solve the problem of meaningless labels. The authors presented a descriptive-based text classification that generates class-specific descriptions using the reinforcement learning approach. Authors in (Chen 2019) developed a framework to analyze the word-level and sentence-level sentiment analysis using reinforcement learning.

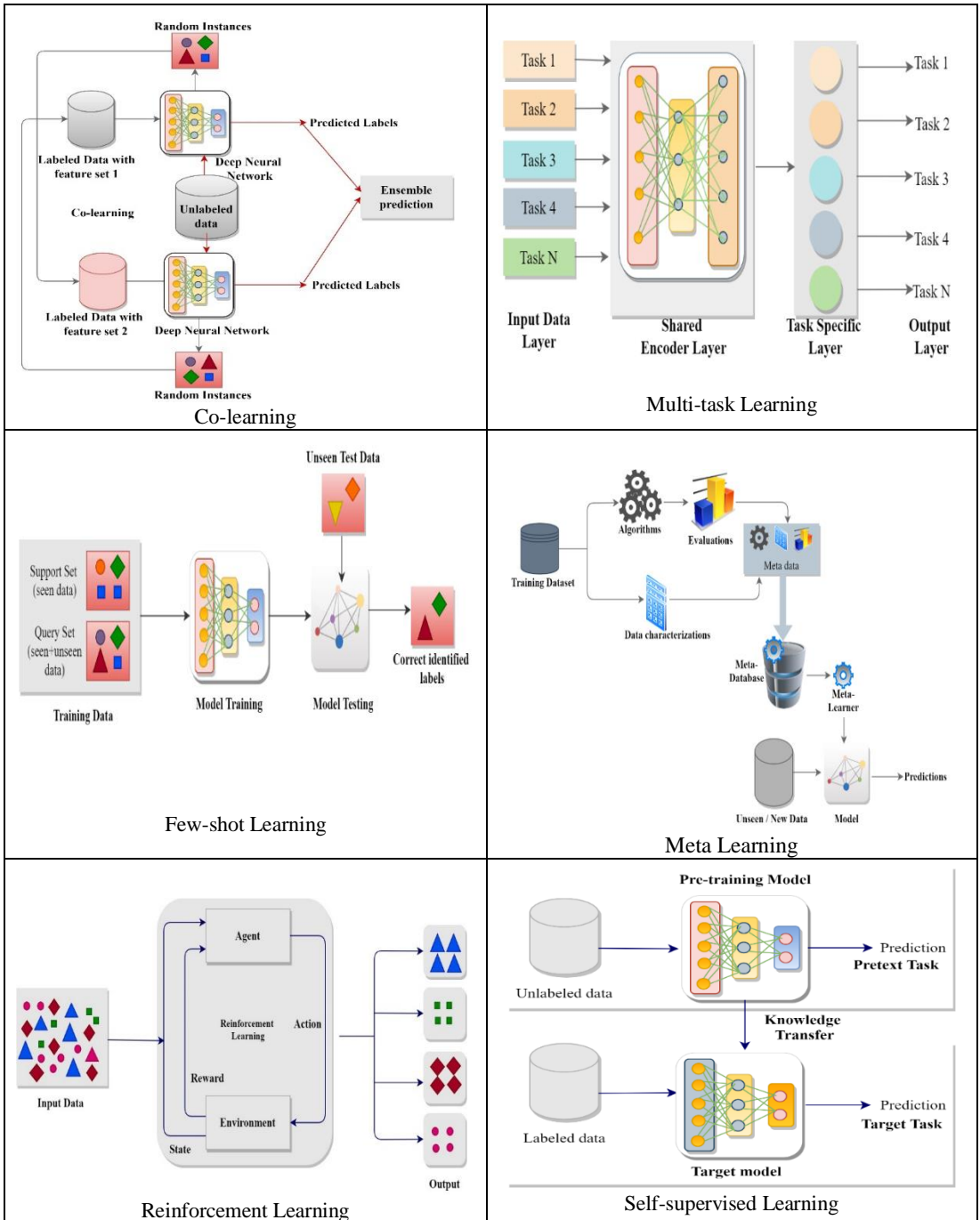

**Fig. 20** Learning Styles

**Table 14** Summary of different learning style

| Ref. | Learning Style | Description | Purpose | Application areas |
|---|---|---|---|---|
| Wu et al. 2020 Helwe et al. 2019 | Co-learning | Multiple classifiers can learn from each other using different features of data | Improve the generalization effectiveness of classifiers using learned features, as well as classification models based on unknown data. | Text classification Sentiment classification |
| Zhang et al. 2021 Akhtar et al. 2020 | Multi-task Learning | Model is trained on multiple tasks simultaneously by sharing representations information instead of training each task separately. | To make a model more robust in a limited dataset. Faster performance. | Stock Prediction Multi-Domain Sentiment Classification, Machine Translation Syntactic Parsing Microblog Analysis |
| Thakkar et al. 2021 Hu et al. 2020 | Zero/Few Shot Learning | It can detect classes that the model has never seen or seen very few classes during training. | To provide good performance in unlabeled datasets. | Sentiment Analysis Multi-Label Text Classification Machine Translation User Intent Classification For Dialog Systems |
| Madrid et al. 2019 Yin et al. 2020 | Meta-Learning | It learns to handle new tasks when tested by performing numerous tasks on training data. | To increase the quantitative performance of models by speeding up the learning process. Allows knowledge transfer. | Text Classification Machine Translation, Dialogue Generation Intent classification Dialog state tracking Question answering |
| Duo et al. 2020 Chen et al. 2019 | Reinforcement Learning | An agent is trained in an interactive environment through a reward and punishment mechanism using feedback from its own actions and experiences. | Used in automated goal-directed learning and sequential decision-making. | Text summarization Question and answering Conversational System Machine Translation |
| Su et al. 2021 Salim et al. 2021 | Self-supervised learning | It trains the system itself by precisely predicting the other part of the data and generating labels using one part of the data. | To improve the performance of classification from unlabeled data and also when a dataset is large to label it. | Classification task Sentiment analysis Text generation |

*6) Self-supervised learning – Self*-supervision is another term for self-supervised learning. In this type of learning, the model learns themselves by utilizing one part to anticipate the other of the data and accurately create annotations; this is called a pretext task and then using representations learned from the pretext task for other supervised tasks. Finally, the self-supervision learning method transforms an unsupervised learning problem into a supervised learning problem. Self-supervised learning is used where the dataset is too large, and labeling/annotating it will be time-consuming. This method makes data labeling an automated process, and human intervention is eliminated. (Su et al. 2021) the authors proposed an approach to automatically train attention supervision information for neural Aspect based sentiment analysis. In (Salim et al. 2021) The authors created SSentiA (Self-supervised Sentiment Analyzer), a self-supervised hybrid methodology for sentiment classification from unlabeled data that combines a machine learning classifier with a lexicon-based strategy.

## 6 CONCLUSION

Digital online media has increased demand in recent years. This article provides a comprehensive review of the literature on TBED using digital online platforms. In many applications, artificial intelligence techniques have been applied to digital online media analysis. This review provides the outcomes of a systematic literature review on TBED methods whereby the authors intended to highlight the employed emotion models, adopted features extraction methods, approaches, the approved data sets, adopted application domains, and the challenges with respect to TBED. In this study, different SLR phases were planned, conducted, and executed on TBED. The research questions were based on criteria like datasets, approaches, application domains, and investigating gaps in the literature. Existing literature shows different artificial intelligent approaches used like deep learning-based, machine learning-based, rule-based, and keyword-based approaches for TBED. Deep learning and machine learning-based approaches are trending with the help of available datasets and automated feature extraction methods. The authors surveyed publicly available datasets for TBED. In RQ 3, different applications domains are explored for TBED research. Research challenges in the field of TBED are identified, such as domain dependency, imbalanced datasets, and future directions for TBED based on artificial intelligence are presented.